\documentclass[journal]{IEEEtran}
\usepackage[T1]{fontenc}% optional T1 font encoding

\usepackage{cite}
\usepackage{amsmath}
\usepackage{amsthm}
\usepackage{amssymb}
\usepackage{dsfont}
\usepackage{graphicx}
\usepackage{multicol}
\usepackage{textcomp}
\usepackage{xcolor}
\usepackage{xspace}
\usepackage{algorithm,algorithmic}
\usepackage{hyperref}

\newtheorem{ass}{Assumption}
\newtheorem{Lemma}{\rm \textbf{Lemma}}
\newtheorem{rem}{Remark}
\usepackage{etoolbox}
\usepackage[normalem]{ulem}
\DeclareMathAlphabet{\mathcal}{OMS}{cmsy}{m}{n}

\makeatletter
\patchcmd{\@makecaption}
{\scshape}
{}
{}
{}
\makeatother

\newtheorem{innercustomthm}{Theorem}
\newenvironment{customthm}[1]
  {\renewcommand\theinnercustomthm{#1}\innercustomthm}
  {\endinnercustomthm}
\usepackage[cmintegrals]{newtxmath}

\makeatletter
\newcommand\fs@betterruled{%
  \def\@fs@cfont{\bfseries}\let\@fs@capt\floatc@ruled
  \def\@fs@pre{\vspace*{5pt}\hrule height.8pt depth0pt \kern2pt}%
  \def\@fs@post{\kern2pt\hrule\relax}%
  \def\@fs@mid{\kern2pt\hrule\kern2pt}%
  \let\@fs@iftopcapt\iftrue}
\floatstyle{betterruled}
\restylefloat{algorithm}
\makeatother

\usepackage{fancyhdr}
\begin{document}

\title{Node Selection Toward Faster Convergence for Federated Learning on Non-IID Data}

\author{Hongda Wu,~\IEEEmembership{Student Member, IEEE} and
        Ping Wang,~\IEEEmembership{Fellow,~IEEE}
        % <-this % stops a space
\thanks{Hongda Wu and Ping Wang are with the Department of Electrical Engineering and Computer Science, Lassonde School of Engineering, York University, Toronto, ON M3J 1P3, Canada (e-mail: hwu1226@eecs.yorku.ca; pingw@yorku.ca)}% <-this % stops a space
\thanks{}% <-this % stops a space
}

\maketitle

\begin{abstract}
Federated Learning (FL) is a distributed learning paradigm that enables a large number of resource-limited nodes to collaboratively train a model without data sharing. The non-independent-and-identically-distributed (non-i.i.d.) data samples invoke discrepancies between the global and local objectives, making the FL model slow to converge. In this paper, we proposed \texttt{Optimal Aggregation} algorithm for better aggregation, which finds out the optimal subset of local updates of participating nodes in each global round, by identifying and excluding the adverse local updates via checking the relationship between the local gradient and the global gradient. Then, we proposed a \texttt{P}robabilistic \texttt{N}ode \texttt{S}election framework (\texttt{FedPNS}) to dynamically change the probability for each node to be selected based on the output of \texttt{Optimal Aggregation}. \texttt{FedPNS} can preferentially select nodes that propel faster model convergence. The convergence rate improvement of \texttt{FedPNS} over the commonly adopted Federated Averaging (\texttt{FedAvg}) algorithm is analyzed theoretically. Experimental results demonstrate the effectiveness of \texttt{FedPNS} in accelerating the FL convergence rate, as compared to  \texttt{FedAvg} with random node selection.\end{abstract}


% \fi
\IEEEpeerreviewmaketitle

\section{Introduction}
\IEEEPARstart{W}{ith} the rapid growth of computational capability at mobile edge sides, next-generation computing network is experiencing a paradigm shift from traditional cloud computing to Mobile Edge Computing (MEC) systems \cite{chiang2016fog}\cite{8436042}. With the deployed computational power and storage capability, MEC systems construct the node-edge-cloud architecture in supporting the applications at resource-constrained nodes that require low latency communication (e.g., autonomous driving) or high throughput (e.g., content delivery network) \cite{7574435}. Edge nodes such as sensors, mobile devices, and connected vehicles are generating an unprecedented amount of data consistently and coupled with cutting-edge Machine Learning (ML) / Deep Learning (DL) techniques, the MEC system is able to conduct intelligent inference (e.g., road congestion prediction \cite{zhang2015testing}) and perceptive control (e.g., unmanned aerial vehicles (UAVs) swarm navigation \cite{8676325}).

In traditional ML fashion, in order to train a complex DL model with millions of model parameters, a tremendous amount of data aggregated from multiple edge nodes is typically needed, which is offloaded via wireless network to edge server. However, collecting data for model training is unrealistic from privacy, security, regulatory, or necessity perspectives. With the ever-increasing computational capability on edge nodes, it becomes more attractive to perform model training on the edge node side instead of sending raw data to the edge server. To this end, Federated Learning (FL) has emerged as a variant of the previous Distributed ML (DML) manner, which decouples the data acquisition and model training at the edge server \cite{McMahan2017CommunicationEfficientLO, konevcny2016federated}. In general, FL systems aim to optimize a global model under the orchestration of an edge server, which allows the collaboration of multiple edge nodes for data augmentation while keeping training data locally. FL involves several communication rounds, each of which includes local model training, model update transmission, and global model aggregation. Along the iterative process, the edge server is able to train a statistical model that is suitable for all participating nodes without accessing user-sensitive data. The improved data confidentiality and reduced volume of communication cost make FL one of the most promising technologies for future network intelligence \cite{8865093}.

Nonetheless, a fundamental challenge for FL in comparison with the optimization in DML, where algorithms run on independent and identically distributed (i.i.d) data samples partitioned from a large dataset, is the data heterogeneity \cite{ li2018federated,zhao2018federated, wang2020optimizing}. To be more specific, the model is updated via feature learning on local data samples, which are user-specific and reveal a different pattern. The data samples across participating nodes may not be independent and identically distributed (non-i.i.d.). Since participating nodes in each iteration are selected randomly, data distribution on nodes cannot represent the global data distribution. Training on nodes with non-i.i.d. datasets will lead to the biased model update, which stagnates model convergence and reduces the model accuracy substantially, and consequently invokes additional communication rounds to resource-constrained edge nodes \cite{zhao2018federated, wang2020optimizing}. Though a relatively small amount of data is sent (i.e., in general, model parameters have a smaller size than the raw training data), communication time in FL is proven to be the critical bottleneck for FL due to the network uncertainty, bandwidth limitation, and straggler effect, etc. \cite{wang2019adaptive}.

In this paper, we design a node selection\footnote{A critical property that differentiates FL from a typical distributed optimization problem is the massively distributed nodes \cite{McMahan2017CommunicationEfficientLO}. Therefore, in each round, a small fraction of nodes is selected for participation.}  scheme to improve the convergence rate of FL with non-i.i.d nodes, called \texttt{FedPNS}, which is a \texttt{P}robabilistic \texttt{N}ode \texttt{S}election framework with contribution-related criteria. We find out global model aggregation over all participating nodes is not of necessity, whereas excluding some adverse local updates may lead to a better global model in terms of model accuracy. In order to improve the expected decrement of FL loss in each round,  we propose an \texttt{Optimal Aggregation} algorithm to determine the optimal subset of local updates (from the participating nodes) for global model aggregation, which utilizes the inner product between local gradients and the global gradient\footnote{We use local/global gradient and local/global update interchangeably.} as the indicator. By applying the result from \texttt{Optimal Aggregation}, the data heterogeneity can be profiled, which is used to adjust the probability for each node to be selected in the subsequent global rounds. Consequently, the server can preferentially select nodes that propel faster model convergence. Note that our probabilistic node selection is conducted on the server-side, which does not impose additional communication costs. Our main
contributions in this paper are as follows

\begin{itemize}
\item We analyze the convergence bound of the commonly adopted Federated Averaging (\texttt{FedAvg}) algorithm \cite{McMahan2017CommunicationEfficientLO} from a theoretical perspective and derive the expected decrease of FL global loss, considering the data heterogeneity and the way to aggregate local updates.  
\item We challenge the necessity of global model aggregation over local updates of all participating nodes and propose \texttt{Optimal Aggregation} to identify and exclude the potential adverse local updates, which enlarges the expected decrease of global loss in each round.
\item We design \texttt{FedPNS}, a \texttt{P}robabilistic \texttt{N}ode \texttt{S}election scheme that enables server to dynamically adjust the probability for each node to be selected in each round, based on the result of \texttt{Optimal Aggregation}. \texttt{FedPNS} tendentiously selects nodes that boost model convergence. The convergence rate improvement of \texttt{FedPNS} over \texttt{FedAvg} is illustrated theoretically and the imposed computational complexity of \texttt{FedPNS} is discussed. 
\item We empirically evaluate the performance of \texttt{FedPNS} via extensive experiments using the synthetic dataset and real datasets with different learning objectives. The experimental results show the effectiveness of \texttt{FedPNS} in improving the convergence rate of the FL model compared
with the commonly adopted \texttt{FedAvg} algorithm.
\end{itemize}

\section{Related Work}
\label{II}

Some existing works on FL focus on the communication cost reduction, with the aim of directly reducing communication cost on the wireless link, where typical methods range from important-based updating \cite{8885054}, \cite{lin2018deep}, model quantization \cite{seide20141}, and analog aggregation \cite{8870236}. In particular, Wang \textit{et al.} \cite{8885054} proposed identifying the irrelevant update at the node side caused by different data distribution. Communication cost is reduced by excluding irrelevant updates from local nodes. Similarly, authors in \cite{lin2018deep} introduced a concept, namely important gradient, where communication reduction is achieved by sending the gradient with a larger magnitude. Different from \cite{8885054, lin2018deep}, Seide \textit{et al.} \cite{seide20141} proposed 1-bit stochastic gradient descent to reduce model transfer data size and achieved 10$\times$ speed up in speech applications. Zhu \textit{et al.} \cite{8870236} proposed to utilize analog aggregation rather than digital aggregation. By exploiting the waveform-superposition property of a multi-access channel, model transmission and aggregation are realized over wireless links simultaneously.

Another series of studies concentrate on the algorithmic perspective via handling the inherent non-i.i.d. data distribution across participating nodes, aiming to reduce communication rounds in FL. These studies include adaptive tuning local training \cite{li2018federated}, weighting design for model aggregation \cite{wu2021fastconvergent}, and node selection strategies \cite{nishio2019client, 9337227, cho2020client, chen2021communication, chen2020convergence,ren2020scheduling,chen2020optimal, rizk2021optimal}. The algorithm \texttt{FedProx} by Li \textit{et al.} \cite{li2018federated} uses a regularization term to balance the optimizing discrepancy between the global and local objectives, and allowing participating nodes to perform a variable number of local updates, to consequently overcome the non-i.i.d. data distribution and resource heterogeneity. Authors in \cite{wu2021fastconvergent} exhibited a contribution-related weighting design, namely \texttt{FedAdp} to boost FL convergence rate in the presence of nodes with non-i.i.d. data samples, which assigns distinguished weights for participating nodes according to the correlation between local objective and global objective revealed by gradient information. 

In general, to avoid long-tail waiting time in synchronous aggregation protocol, FL algorithm randomly selects a subset of nodes (i.e., partial node participation) in each round to participate in local training (e.g., \texttt{FedAvg} \cite{McMahan2017CommunicationEfficientLO}, \texttt{FedProx} \cite{li2018federated}, \texttt{CMFL} \cite{8885054}). Compared with DML, candidate nodes in FL are more heterogeneous with regard to computation/communication capability, wireless connection, and data quality. Therefore, a carefully designed node selection is beneficial for performance improvement. Several works are carried out focusing on node selection design to improve the FL convergence rate, taking the system heterogeneity and uncertainty of wireless medium into consideration \cite{nishio2019client, 9337227, chen2021communication, chen2020convergence,ren2020scheduling}. 
 Specifically, Nishio \textit{et al.} \cite{nishio2019client} proposed to select nodes intentionally based on the resource condition on nodes. Amiria \textit{et al.} \cite{9337227} designed a node scheduling algorithm by considering the significance of local update measured by $\ell2$-norm and channel condition separately or jointly. For example, in \texttt{BN2} algorithm \cite{9337227}, the server first selects a macro set of nodes to participate in local training.  Then a subset of the macro set is finally chosen for model aggregation by ordering the norm of gradient transmitted from nodes of the macro set. In \cite{cho2020client}, the authors proposed biased client selection strategies, that is, preferentially choosing the node with higher local loss. Though the contribution-related loss measurement leads to a faster convergence, the selection skewness imposes potential error, and the local loss measurement results in additional communication and computation cost. Differently, references \cite{chen2021communication, chen2020convergence, ren2020scheduling, chen2020optimal, rizk2021optimal} focus on probabilistic node selection strategy where each node is eligible to contribute to the global model. In particular, Chen \textit{et al.} \cite{chen2021communication} considered the limited bandwidth resource for model transmission where node selection for global model aggregation is of importance. The proposed method measures the node contribution according to the norm of local updates, by which the probability for each node to be selected is calculated so as to execute the node selection procedure. The nodes with higher norm of local updates are chosen with higher probability, thus boosting the convergence rate when limited bandwidth resource is provided. Along with \cite{chen2021communication}, authors in \cite{chen2020convergence} proposed to use Artificial Neural Networks (ANNs) as a predictor to estimate the model updates of nodes that are not allocated the bandwidth for transmission, based on the model updates that are successfully transmitted using limited bandwidth resource. The additionally included model updates further accelerate the model convergence. Authors in \cite{ren2020scheduling}  proposed a probabilistic design by considering the importance of local update and transmission latency, where the importance of local update is evaluated by gradient divergence between local gradients and the ground truth global gradient. The probability for node selection is finally determined by the local gradient norm and transmission latency.  Chen \textit{et al.} \cite{chen2020optimal} designed an importance sampling scheme that selects more informative nodes. The node sampling procedure minimizes the variance of local gradients for aggregation, while the probability for each node to be chosen is proportional to the norm of local updates. In addition, authors in \cite{rizk2021optimal} applied importance sampling for node selection on the server level and data selection on the node level. Similar to \cite{chen2020optimal}, the optimal node selection is achieved by minimizing the bound on the variance of gradient noise, i.e., the estimation error of the global gradient because of the partial node participation. The probability for each node to be chosen is proportional to the norm of its local updates. 
 
None of the aforementioned node selection designs analyzed the impact of data heterogeneity on node selection. Given the heterogeneous training samples across nodes, the magnitude of local gradient norm is deficient in reflecting the contribution from each of those nodes, which is empirically shown in Section V-D, since local gradients may not align with the global gradient.
In contrast to the above research, our work in this paper builds on the data heterogeneity perspective and designs a probabilistic model to choose participating nodes. The proposed method scrutinizes the relationship between local gradients and the global gradient so as to adjust the probability for each node to be selected, which is different from the criteria (i.e., the norm of local gradient/update) adopted in \cite{9337227, chen2021communication, chen2020convergence, ren2020scheduling, chen2020optimal, rizk2021optimal}. Since FedAvg algorithm may struggle to converge on non-i.i.d. data, it is not trivial to profile the distribution of data samples across nodes. Upon identifying the contribution difference of nodes, it is profitable to accelerate model convergence by choosing nodes tendentiously, as compared with random selection. 

The rest of the paper is organized as follows. Section \ref{III} provides the preliminary and implementation of federated learning and the challenge of non-i.i.d. data on FL. In Section \ref{IV}, the convergence analysis, the proposed aggregation scheme and probabilistic node selection, and complexity analysis are presented. Experimental results are shown in Section \ref{V}, and the conclusion is presented in Section \ref{VI}.

\section{Preliminaries}
\label{III}
In this section, we first introduce the key ingredients behind federated learning, including the system model (Section \ref{III-A}) and the practical algorithm design to solve federated learning problem (Section \ref{III-B}). Then, the challenge of FL on heterogeneous data is analyzed (Section \ref{III-C}).

\subsection{Federated Learning Model}
\label{III-A}
In general, federated learning methods \cite{McMahan2017CommunicationEfficientLO, li2018federated, Li2020On}, are designed to handle the consensus learning task in a decentralized manner, where a central server coordinates the global learning objective and multiple devices train the model with locally collected data. Consider a network with $\mathcal{K}$ local nodes (i.e., $i \in \{1,2,\cdots, |\mathcal{K}| \}$), where each node $i$ possesses a local (private) dataset $\mathcal{D}_i$ with size $D_i$. 
The nodes are connected with a central server and seek to collaboratively find a global model parameterized by $\mathbf{w}$ that minimizes the empirical risk
\begin{align*}
\label{eq1}
F( \mathbf{w} )  = \frac{1}{\sum _{i=1}^{|\mathcal{K}|} D_i} \sum _{i=1}^{|\mathcal{K}|} \sum _{\{\mathbf {x}, y\} \in \mathcal{D}_i}  f(\mathbf{w}, \mathbf {x}, y) , \tag{1}
\end{align*}
where $f(\mathbf{w}, \mathbf {x}, y)$ is the composite loss for training sample $\{\mathbf {x}, y\}$. Specifically, in the context of $C$-class classification problem hereinafter, each training sample $\{\mathbf {x}, y\} \in \mathcal{D}_i$ is assumed to contain a feature vector $\mathbf {x}$ and label $y$ over feature space $\mathbb{X}$ and label space $\mathbb{Y}$ (i.e., $\mathbb{Y} = [C]$, where $[C] = \{1, \cdots, C \}$). For each available training sample $\{\mathbf {x}, y\} \in \bigcup_i\mathcal{D}_i$ in FL problem, 
the federated learning model parameterized by $\mathbf{w}$ is considered to learn the predicted probability vector $\bar {\mathbf {y}}$, i.e., $ \bar {\mathbf {y}}|\sum_{j=1}^C \bar y_j =1, \bar y_j \geq 0, \forall j \in [C]$, with empirical risk.

From a federation perspective, the global objective $F( \mathbf{w} )$ in (\ref{eq1}) is surrogated by local objectives $F_{i}(\mathbf {w})$ and can be further represented as follows
\begin{align*}
\label{eq2}
F( \mathbf{w} )  =  \sum _{i=1}^{|\mathcal{K}|} \frac{D_i}{\sum_{i=1}^{|\mathcal{K}|} D_i} F_{i}(\mathbf {w}), \tag{2}
\end{align*}

For node $i \in \mathcal{K}$, $F_i( \mathbf{w} )$ commonly measures the local empirical risk (e.g., cross entropy loss) over the dataset $\mathcal{D}_i$ with possibly differing data distribution $q^{(i)}$, which is defined as follows
\begin{align*}
\label{eq3}
 F_i( \mathbf{w} ) & = 
\mathbb{E}_{\mathbf{x},y \backsim q^{(i)}}  \left[- \sum_{j=1}^C {\mathds{1}}_{ y=j} {\rm log}l_j (\mathbf{w}, \mathbf {x}, y)  \right] \\
& = - \sum_{j=1}^C q^{(i)} (y=j) \mathbb{E}_{\mathbf{x}| y=j} \left[ {\rm log}l_j (\mathbf{w}, \mathbf {x}, y) \right], \tag{3}
\end{align*}
where $l_j (\mathbf{w}, \mathbf {x}, y)$ denotes the probability that the data sample $\{\mathbf{x}, y\}$ is classified as the $j$-th class given model $\mathbf{w}$. $q^{(i)} (y=j)$ denotes the data distribution on node $i$ over class $j \in [C]$.

\subsection{\texttt{FedAvg} with Partial Node Participation}
\label{III-B}
The most commonly used algorithm to solve (\ref{eq2}) is Federated Averaging (\texttt{FedAvg})  \cite{McMahan2017CommunicationEfficientLO, Li2020On}, where the training consists of multiple communication rounds. At each communication round $t$, the server selectes a fraction $c$ of nodes $|\mathcal{S}_t| = c|\mathcal{K}|$ to participate in the training. Taking the global model $\mathbf{w}^{t-1}$ in previous round as the reference, each participating node $i\in \mathcal{S}_t$ performs local Stochastic Gradient Descent (SGD) to optimize its objective
\begin{align*}
\label{eq4}  
 \mathbf{w}_i^t  =  \mathbf{w}^{t-1} - \eta \nabla F_i(\mathbf{w}^{t-1}) 
 ,\tag{4}
\end{align*}
where $\eta$ is the learning rate and $\nabla  F_i(\cdot)$ is the gradient\footnote{Through this paper, the gradient refers to the stochastic version instead of the actual gradient calculated from the entire dataset.} at node $i$. (\ref{eq4}) gives a general principle of SGD optimization, where $\mathbf{w}_i^t$ is the result after $\tau$ local updates of mini-batch SGD (i.e., $\tau = \frac{D_i}{B}E$,  where $E$ is the number of local training epochs, $B$ is the batch size of mini-batch training samples).  

The participating nodes then communicate their model update $\Delta_i^t = \mathbf{w}_i^t  - \mathbf{w}^{t-1} $  back to the server, which aggregates them and updates the global model\footnote{It is worth to mention that the aggregation scheme is applied over all nodes in vanilla \texttt{FedAvg}\cite{McMahan2017CommunicationEfficientLO}, i.e., $\Delta^t =  \sum _{i \in  \mathcal{S}_t} \psi_i\Delta_i^t + \sum _{i \in \mathcal{K}- \mathcal{S}_t} \psi_i \mathbf{w}^{t-1} $, {where $\psi_i = \frac{D_i}{\sum_{i=1}^{|\mathcal{K}|} D_i}$.}  The subsequent work \cite{li2018federated} proposed a variant of aggregation over participating nodes as in (\ref{eq5}). Hereinafter, \texttt{FedAvg} denotes the algorithm that involves random selection and partial aggregation of nodes with equal data size\cite{li2018federated}.} as follows
\begin{align*}
\label{eq5}
\Delta^t  &= \frac{1}{|\mathcal{S}_t|} \sum _{i \in  \mathcal{S}_t}  \Delta_i^t \\
  \mathbf{w}^t &= \mathbf{w}^{t-1} + \Delta^t.
  \tag{5}
\end{align*}

\subsection{The Challenges of Non-i.i.d. Data Distribution}
\label{III-C}
Though \texttt{FedAvg} can achieve a decent convergence rate with random node selection policy and simple averaging design, partial node participation and non-i.i.d. training data slows the convergence rate \cite{Li2020On}, which is also observed in \cite{zhao2018federated, wang2020optimizing, wang2019adaptive, cho2020client, wu2021fastconvergent}. 
Since communication cost becomes a critical bottleneck in FL, one can increase the local computing (i.e., more local updates), which is shown to be beneficial to save communication rounds and improve the convergence rate \cite{McMahan2017CommunicationEfficientLO, wang2019adaptive}.

However, model performance on non-i.i.d. dataset is not satisfactory, even with increased local computing \cite{ Li2020On, stich2018local}. This is because the local objective  $F_{i}(\mathbf {w})$, which the local optimizer minimizes, is closely related to data distribution $q^{(i)}$. In trivial node selection policy (e.g., random selection in \texttt{FedAvg}), the distribution of data samples on selected nodes differs from each other. Local updates lead the model towards optima to its local objective, which is deviated from the global objective in a non-i.i.d. setting, causing training instability that makes the FL model struggle to converge.

It is crucial to understand and analyze the non-trivial node selection policy from the data heterogeneity perspective, identifying and choosing the nodes that contribute better to model convergence. By taking the inner product between the local update and global update as the criterion, which implicitly profiles the difference between data distribution on nodes and population distribution, we first identify the nodes whose updates adversely contribute to the global update. By excluding the potential adverse local updates and reducing the probability for those nodes to be selected, one can ensure that the node with a higher contribution to the decrease of global loss enjoys a higher probability of being chosen. Consequently, the non-trivial node selection accelerates model convergence compared with \texttt{FedAvg}.

\section{Contribution-based Node Selection}
\label{IV}
In this section, we design a probabilistic node selection scheme to improve the convergence rate of federated learning. For FL with the heterogeneous dataset, we analyze the convergence property of \texttt{FedAvg}  theoretically (Section \ref{IV-A}). In Section \ref{IV-B}, we challenge the necessity of global model aggregation over all participating nodes. Then, the \texttt{Optimal Aggregation} algorithm is proposed, which can identify and exclude the adverse local updates to make greater progress on reducing the expected decrement of global loss in each round. The FL with Probabilistic Node Selection (\texttt{FedPNS}) is proposed based on the result of \texttt{Optimal Aggregation}. \texttt{FedPNS} adjusts the probability for each node to be selected, and the server is able to preferentially select nodes that propel a faster model convergence (Section \ref{IV-C}). The convergence rate improvement of \texttt{FedPNS} over \texttt{FedAvg} is analyzed theoretically (Section \ref{IV-D}) and the computation complexity of \texttt{FedPNS} is discussed in Section \ref{IV-E}.

\subsection{Convergence Analysis}
\label{IV-A}
For theoretical analysis purposes, we employ the following assumptions to the loss function, which have also been commonly made in the literature \cite{li2018federated, Li2020On, nguyen2020fastconvergent, stich2018local}.

\begin{ass} 
\label{ass1}
\textbf{ Convex, $\zeta$-Lipschitz, and $L$-smooth}.\\
$ F_i( \mathbf{w})$ is convex, $\zeta$-Lipschitz, and $L$-smooth for all node $i$, \\
 i.e., $\Vert F_i( \mathbf{w}) - F_i( \mathbf{w'}) \Vert  \leq \zeta \Vert  \mathbf{w} -  \mathbf{w'} \Vert $, \\ $ \quad \Vert \nabla F_i( \mathbf{w}) -\nabla F_i( \mathbf{w'}) \Vert  \leq L \Vert  \mathbf{w} -  \mathbf{w'} \Vert $, 
 for any $ \mathbf{w}$, $ \mathbf{w'}$.
\end{ass}

 Based on Assumption \ref{ass1}, the definition of $F(\mathbf{w}) $, and triangle inequality, we can easily get that $F( \mathbf{w}) $ is convex, $\zeta$-Lipschitz, and $L$-smooth.
\begin{ass} 
\label{ass2}
\textbf{$\delta$-local dissimilarity}.\\
Local loss functions $ F_i(\mathbf{w}^{t})$ are $\delta$-local dissimilar at $\mathbf{w}^{t}$, \\ i.e., $\mathbb{E}_{i \backsim \mathcal{S}_t}\left[ \Vert  \nabla F_i( \mathbf{w}^{t}) \Vert^2  \right] \leq \Vert \nabla F( \mathbf{w}^{t}) \Vert^2 \delta^2$ for $i \in \mathcal{S}_t$ and $t = 1, \cdots, T$, where $T$ is the number of global rounds. $\mathbb{E}_{i \backsim \mathcal{S}_t}[\cdot]$ denotes the expectation over participating nodes $\mathcal{S}_t$ with weight $\frac{1}{|\mathcal{S}_t|}$ (as in (\ref{eq5})).  $\nabla F(\mathbf{w}^{t})$ is the global gradient at the $t$-th global round defined as $\nabla F(\mathbf{w}^{t}) = \frac{1}{|\mathcal{S}_t|} \sum_{i\in \mathcal{S}_t} \nabla F_i(\mathbf{w}^{t})$.
\end{ass}
\begin{ass} 
\label{ass3}
\textbf{Bounded gradient}.\\
The norm of gradient in each node is bounded, i.e., $\Vert \nabla F_i(\mathbf{w}^t) \Vert \leq \gamma_i$ for all $i \in \mathcal{K}$ and $t = 1, \cdots, T$.
\end{ass}

Assumption \ref{ass1} is standard, which can be satisfied when the logistic regression with cross entropy loss is adopted \footnote{More examples include $\ell2$-norm regularized linear regression with mean square error, and the support vector machine with hinge loss.}. The discrepancy between the local objective and global objective caused by the data heterogeneity is captured by Assumption \ref{ass2}, which has been made in previous work \cite{Li2020On, nguyen2020fastconvergent}. 
As the data distribution across participating nodes becomes more heterogeneous, the local updates (i.e., gradient) will diverge from each other, and $\delta$ will increase. On the other hand, if the data samples on participating nodes follow the same data distribution,
the local gradients become more similar and $\delta$ goes to 1. Assumption \ref{ass3} has been made in different forms by previous works \cite{Li2020On, stich2018local}. Besides, with $\mathbf{w}$ trained by heterogeneous data, $\gamma$ is different for different nodes, which is closely related to the data distribution on each node. If the data distribution on node $i$ is more similar to the population distribution over all nodes, $\gamma_i$ is lower, and vice versa. This observation is empirically illustrated in Section \ref{V-D}.

\begin{Lemma}
\label{thm1}
Let assumptions \ref{ass1} and \ref{ass2} hold. Suppose that  $\mathbf{w}^t$ is not a stationary solution, the expected decrement on the global loss of FedAvg between two consecutive rounds satisfies
\begin{align*}
\label{eq6}
F(\mathbf{w}^{t+1})  & \leq  F(\mathbf{w}^{t}) -\eta \mathbb{E}_{i \backsim \mathcal{S}_t}\left[     \langle \nabla F(\mathbf{w}^{t}),  \nabla F_i(\mathbf{w}^{t})  \rangle \right] \\ & +  \frac{L\eta^2}{2}  \Vert \nabla F( \mathbf{w}^{t}) \Vert^2 \delta^2,
\tag{6}
\end{align*}
where $\eta$ is the learning rate of SGD, $\langle \cdot \rangle$ is the inner product operation, and $\Vert \cdot \Vert$ denotes the $\ell2$-norm of a vector.
\end{Lemma}

The proof of {Lemma} \ref{thm1} is presented in Appendix\ref{apex1}. Lemma \ref{thm1} provides a bound on how rapid the decrease of the global FL loss can be expected. The decrease of global FL loss between two consecutive rounds shows a dependency on $\delta$, which represents the variance between local data distributions, and the aggregation strategy $\mathbb{E}_{i \backsim \mathcal{S}_t} [\cdot]$, where $\nabla F(\mathbf{w}^{t})$ is obtained by aggregating over local updates from all participating nodes, i.e., $\nabla F_i(\mathbf{w}^{t}), i \in \mathcal{S}_t$ with weight $1/|\mathcal{S}_t|$. 

\subsection{Aggregation with Gradient Information}
\label{IV-B}

In vanilla \texttt{FedAvg} \cite{McMahan2017CommunicationEfficientLO} and the subsequent work \cite{li2018federated, wang2019adaptive, zhao2018federated, wang2020optimizing}, the averaging technique is used for global update aggregation due to its simplicity. One can challenge the inherent rule that the global update is aggregated over local updates of all participating nodes since the local updates may contribute global model in an adverse way. As a sanity check, at any communication round $t$, the local updates from the participating nodes whose inner product between their gradients and the global gradient is negative i.e., $\langle \nabla F(\mathbf{w}^{t}),  \nabla F_i(\mathbf{w}^{t})  \rangle < 0$, will slow the model convergence because of the reduced expected loss decrement (i.e., a lower expectation value as in (\ref{eq6})) in this round.  As such, it is not trivial to exclude the adverse local updates, which is realized by examining the value of expectation term in Lemma \ref{thm1}, as illustrated later. Excluding adverse local updates gives an impact on the reduction of overall data heterogeneity, which, in the meanwhile, changes the relationship between local gradients and the global gradient $\langle \nabla \bar{F}(\mathbf{w}^{t}),  \nabla F_i(\mathbf{w}^{t}) \rangle $, where $\nabla \bar{F}(\mathbf{w}^{t}) = \frac{1}{|\bar{ \mathcal{S}_t}|} \sum_{i\in \mathcal{S}^*_t } \nabla F_i(\mathbf{w}^{t})$ is defined over $\mathcal{S}^*_t$, i.e., the subset of participating nodes $\mathcal{S}_t$ after successfully excluding the nodes with adverse local updates.

To find the optimal subset of local updates to aggregate, we first check the expectation term $\mathbb{E}_{i \backsim \mathcal{S}_t}\left[  \langle \nabla F(\mathbf{w}^{t}),  \nabla F_i(\mathbf{w}^{t})  \rangle \right]$ in Lemma \ref{thm1} and exclude the local updates from participating nodes $k$, i.e., $k \in \mathcal{S}_t - \bar{\mathcal{S}_t}$ if $\mathbb{E}_{i \backsim \bar{\mathcal{S}_t} }\left[  \langle \nabla \bar{F}(\mathbf{w}^{t}),  \nabla F_i(\mathbf{w}^{t})  \rangle \right] > \mathbb{E}_{i \backsim  \mathcal{S}_t}\left[  \langle \nabla F(\mathbf{w}^{t}),  \nabla F_i(\mathbf{w}^{t})  \rangle \right]$ is satisfied. However, excluding local updates gives an impact on the global update and overall data heterogeneity, i.e., $\Vert \nabla F( \mathbf{w}^{t}) \Vert^2 \delta^2$, the last term on the right hand side of (\ref{eq6}), which makes the expected decrement of global loss, i.e., $\Delta F( \mathbf{w}^{t}) = \frac{L\eta^2}{2}  \Vert \nabla F( \mathbf{w}^{t}) \Vert^2 \delta^2 -\eta \mathbb{E}_{i \backsim \mathcal{S}_t}\left[\langle \nabla F(\mathbf{w}^{t}),  \nabla F_i(\mathbf{w}^{t})  \rangle \right]$, difficult to be analyzed quantitatively given $L$ and $\delta$. Therefore, in the second step, test loss is adopted to ensure that excluding local updates makes global update better in terms of model convergence, as in \cite{cho2020client}. In particular, the global model $\mathbf{w}^{t+1}$ and $\bar{\mathbf{w}}^{t+1}$ generated by $\nabla F_i(\mathbf{w}^{t}), i \in \mathcal{S}_t$ and $\nabla F_i(\mathbf{w}^{t}), i \in \bar{\mathcal{S}_t}$, respectively, are evaluated using mini-batch of samples with size $\bar B$ that are sampled uniformly at random from $\mathcal{D}_{test}$ (e.g., test dataset in MNIST). 
\begin{algorithm}[t]
\caption{Optimal Local Updates for Aggregation}
 \label{algorithm1}
 \begin{algorithmic}[1]
  \renewcommand{\algorithmicrequire}{\textbf{Procedure} \	\textsc{Optimal Aggregation}}
 \renewcommand{\algorithmicensure}{\textbf{Input:} $\mathcal{S}_t$, $\Delta_i^t$, $v$, temp = $\{\}$ }
 \REQUIRE 
 \ENSURE 
 \STATE $\nabla F(\mathbf{w}_i^{t}) = -\Delta_i^t/\eta $
 \STATE  $\nabla F(\mathbf{w}^{t}) = \frac{1}{|\bar{ \mathcal{S}_t}|} \sum_{i\in  \mathcal{S}_t}  \nabla F_i(\mathbf{w}^{t})$
  \STATE  max = $\mathbb{E}_{i \backsim \mathcal{S}_t}\left[  \langle \nabla F(\mathbf{w}^{t}),  \nabla F_i(\mathbf{w}^{t})  \rangle \right]$\\
  
 \STATE  \textbf{while} $|\mathcal{S}_t| \geq v$ \textbf{do}\\
 
  \STATE \hspace*{1em} temp $\leftarrow$ \textsc{Check Expectation} ($\nabla F_i(\mathbf{w}^{t})$, $\mathcal{S}_t$, temp)

\STATE \hspace*{1em} \textbf{if} \texttt{max}(temp).value $<$ max \textbf{do} \\

\STATE \hspace*{2em} \textbf{break} with $\mathcal{S}^*_t = \mathcal{S}_t$ \\

\STATE \hspace*{1em} \textbf{else} \\

\STATE \hspace*{2em} key = \texttt{max}(temp).key  \\

\STATE \hspace*{2em} ls($\mathbf{w}$), ls($\bar{\mathbf{w}}$), $\bar{ \mathcal{S}_t}$ $\leftarrow$ \textsc{Check Loss} ($\nabla F_i(\mathbf{w}^{t})$, $\mathcal{S}_t$, key)  \\
     
\STATE \hspace*{2em} \textbf{if} ls($\mathbf{w}$) $>$ ls($\bar{\mathbf{w}}$) \textbf{do}   \\
     
\STATE \hspace*{3em} \textbf{break} with $\bar{ \mathcal{S}_t}$, $\mathcal{S}^*_t = \mathcal{S}_t$\\ 

\STATE \hspace*{2em} \textbf{else} \\
\STATE \hspace*{3em} $\mathcal{S}_t, \mathcal{S}^*_t \leftarrow \mathcal{S}_t$.\texttt{pop}(key)

\STATE \hspace*{3em} max $\leftarrow$ temp(key).value 

\STATE \textbf{return} $\mathcal{S}^*_t$, $\bar{ \mathcal{S}_t}$  \\

\STATE  $\mathbf{w}^{t+1}$ $\leftarrow$ \textsc{Global Update } ($\nabla F_i(\mathbf{w}^{t})$, $\mathcal{S}^*_t$)\\
         
    \renewcommand{\algorithmicrequire}{\textbf{Procedure} \	\textsc{Check Expectation}}
 \renewcommand{\algorithmicensure}{\textbf{Input:} $\nabla F_i(\mathbf{w}^{t})$, $\mathcal{S}_t$, temp}
 \REQUIRE 
 \ENSURE
 \STATE  \textbf{for} $i = 1, \cdots, |\mathcal{S}_t| $ \textbf{do} \\ 
   \STATE  \hspace*{1em} $\bar{ \mathcal{S}_t} \leftarrow \mathcal{S}_t$.\texttt{pop}($\mathcal{S}_t[i]$)
\STATE  \hspace*{1em} $\nabla \bar{F}(\mathbf{w}^{t}) = \frac{1}{|\bar{ \mathcal{S}_t}|} \sum_{i\in \bar{ \mathcal{S}_t}}  \nabla F_i(\mathbf{w}^{t})$

  \STATE  \hspace*{1em} temp($\mathcal{S}_t[i]$) = $\mathbb{E}_{i \backsim \bar{ \mathcal{S}_t}}\left[  \langle \nabla \bar{F}(\mathbf{w}^{t}),  \nabla F_i(\mathbf{w}^{t})  \rangle \right]$
  
    \renewcommand{\algorithmicrequire}{\textbf{Procedure} \	\textsc{Check Loss}}
 \renewcommand{\algorithmicensure}{\textbf{Input:} $\nabla F_i(\mathbf{w}^{t})$, $\mathcal{S}_t$, key}
 \REQUIRE 
 \ENSURE
  \STATE $\bar{ \mathcal{S}_t} \leftarrow \mathcal{S}_t$.\texttt{pop}(key)
  \STATE Generate global model $\mathbf{w}^{t+1}$ by $\nabla F_i(\mathbf{w}^{t}), i \in \mathcal{S}_t$
  and $\bar{\mathbf{w}}^{t+1}$ by $\nabla F_i(\mathbf{w}^{t}), i \in \bar{\mathcal{S}_t}$, respectively \\
  \STATE Evaluate $\mathbf{w}^{t+1}$, $\bar{\mathbf{w}}^{t+1}$ by using batch samples (with size $ \bar B$) from $\mathcal{D}_{test}$ and get the loss ls($\mathbf{w}$) and ls($\bar{\mathbf{w}}$), respectively
  \STATE \textbf{return} ls($\mathbf{w}$), ls($\bar{\mathbf{w}}$), $\bar{ \mathcal{S}_t}$
  
   \renewcommand{\algorithmicrequire}{\textbf{Procedure} \	\textsc{Global Update}}
 \renewcommand{\algorithmicensure}{\textbf{Input:} $\nabla F_i(\mathbf{w}^{t})$, $\mathcal{S}^*_t$}
 \REQUIRE 
 \ENSURE
  \STATE Generate $\mathbf{w}^{t+1}$ by $\nabla F_i(\mathbf{w}^{t}), i \in \mathcal{S}^*_t$ via (\ref{eq4}) and (\ref{eq5})
  \STATE \textbf{return} $\mathbf{w}^{t+1}$
  
 \end{algorithmic} 
 \end{algorithm}
 
An iterative algorithm called \texttt{Optimal Aggregation} is proposed for a better local update aggregation in each round, which finds the \textit{optimal} subset of local updates $\Delta_i, i \in \mathcal{S}_t^* \subseteq \mathcal{S}_t$ by excluding the adverse local updates $\Delta_k$,  $k \in \mathcal{S}_t - \mathcal{S}_t^*  $, as in Algorithms \ref{algorithm1}. Specifically, for a given set of participating nodes $\mathcal{S}_t$ in each global round $t$,  the server iteratively removes one of the local updates $\nabla F_i(\mathbf{w}^{t}), i \in \mathcal{S}_t$, generates the potential global gradient, and calculates the expectation term in (\ref{eq6}) (i.e., \textsc{Check Expectation}, line 18-21). If excluding one local update gives a higher expectation value, compared with the case that includes all local updates retained in $\mathcal{S}_t$, 
that local update will be labeled, and loss comparison will be performed to check the loss criterion (\textsc{Check Loss}, line 22-25), otherwise the server keeps all local updates (line 6). If the loss criterion is satisfied (line 13), the labeled local update is eventually removed from set $\mathcal{S}_t$ (line 14). Otherwise, the server keeps that local update retained in $\mathcal{S}_t$ (line 12). The process repeats until
no adverse local update can be found or the number of remaining local updates is below a threshold $v$ (line 4). In Algorithm \ref{algorithm1}, the function \texttt{pop} is defined as removing element (line 14).  The introduced $``$temp$"$ is a dictionary with key-value pairs (line 5) and the function \texttt{max} returns the maximum value (line 6) or the key (i.e., the node index $i$) corresponding to that value (line 9), respectively.

Given a set of participating nodes $\mathcal{S}_t$, the benefits of finding optimal local updates are twofold: (i) Excluding the potential local updates that contribute to the global model adversely results in a larger decrement of the expected loss in each round. (ii) By \textsc{Check Expectation}, the potential adverse nodes $k, k \in \mathcal{S}_t - \bar{\mathcal{S}_t}$ (nodes with non-i.i.d. dataset normally) are identified. This identification can be used for consequent probabilistic node selection, as illustrated in Section \ref{IV-C}. 

\subsection{FL with Probabilistic Node Selection (\texttt{FedPNS})}
\label{IV-C}
Providing the variety of different nodes on contributing global model, to improve the convergence rate, one can seek to preferentially select the nodes with higher contribution (i.e., the nodes with i.i.d. dataset, as observed in \cite{McMahan2017CommunicationEfficientLO, wu2021fastconvergent}). As such, we propose a probabilistic node selection design that dynamically changes the probability for each node to be selected in each communication round, based on their data distribution-related contribution, which can be distinguished by the procedure \textsc{check expectation} in \texttt{Optimal Aggregation}. 

As we know in each round of FL, a number of nodes are selected to participate in the local training and global aggregation. It is natural to lower the node selection probabilities for those nodes whose local updates slow model convergence. Therefore, on the server-side, we propose to dynamically change the probability for each node to be selected via using the output of \texttt{Optimal Aggregation} (i.e., $\bar{ \mathcal{S}_t}$). In particular, the probabilities for those nodes that are labeled by the procedure \textsc{check expectation} (i.e., $i \in \mathcal{S}_t - \bar{ \mathcal{S}_t}$) are decreased according to the parameter $x$ in (\ref{eq7}), and the probabilities for all the rest nodes will be increased.
 \begin{align*}
\label{eq7}
\Delta p_i^t = p_i^t  \cdot  \texttt{min} [(x + \beta)^{\alpha}, \: 1], \quad  i \in \mathcal{S}_t - \bar{ \mathcal{S}_t},  \tag{7}
\end{align*}
where $p_i^t$ and $\Delta p_i^t$ denote the probability for node $i$ to be selected in the $t$-th global round, and its probability decrement in next round, respectively. \texttt{min} function returns the minimum value among all arguments, $x\in (0,1]$ is defined as the ratio between the accumulated times that a node is labeled by the procedure  \textsc{Check Expectation} and the accumulated times that the node is selected, $\alpha \in \mathbb{Z}^+, \beta\in [0,1]$ are coefficients as explained in the following
\begin{itemize}
\item $\lim_{x \to \epsilon} (x + \beta)^{\alpha} \approx 1$, where $\epsilon \varpropto \alpha $ is constant.
\item $\lim_{0 \to x \to \upsilon} (x + \beta)^{\alpha} \approx \beta $, where $\upsilon \varpropto \alpha$ is a constant.
\end{itemize}

$\alpha$ controls how big the probability decrement is achieved by $(x + \beta)^{\alpha}$ given a ratio $x$. For example, a large value of $\alpha$ brings an aggressive decrement since the probability decrement happens in a wide range $(\beta, 1)$ as $x$ increases within a small range $(\upsilon, \epsilon)$, making 
the node selection probability drop very quickly when $x$ grows. Meanwhile, the large $\alpha$ makes node selection sensitive to the identification mistake, which may prevent i.i.d. nodes from being selected in the subsequent rounds. However, setting a small value of $\alpha$ is not consistently effective to differentiate the nodes since the probability change is marginal. $\beta$ is adopted to keep the rate of probability change in a visible range $[\beta, 1]$. From experiments, we find out $\alpha = 2, \beta=0.7$ is a good choice that balances the tradeoff. The choice of $\alpha$ and $\beta$ is empirically investigated in Section \ref{V-C}.

\begin{algorithm}[t]
\caption{FL with Probabilistic Node Selection}
 \label{algorithm2}
 \begin{algorithmic}[1]
  \renewcommand{\algorithmicrequire}{\textbf{Procedure} \	\textsc{Federated Optimization}}
 \renewcommand{\algorithmicensure}{\textbf{Input:} $E, B, \eta, \mathcal{K}, T, p_i^t \, i = 1, \cdots, |\mathcal{K}|$ }
 \REQUIRE 
 \ENSURE 
 \STATE  Server initializes  $\mathbf{w}^{0}, p_i^0 = 1/|\mathcal{K}|$
 
 \STATE \textbf{for} $t = 1, \cdots, T$ \textbf{do} 
   \STATE \hspace*{1em} Server samples a subset $\mathcal{S}_t$ of nodes according to $p_i^{t-1}$
 \STATE \hspace*{1em} Server sends $\mathbf{w}^{t}$ to nodes $i \in \mathcal{S}_t$ \\
 
 \STATE \hspace*{1em} Each node $i \in \mathcal{S}_t$ finds $\mathbf{w}_i^{t}$ to optimize $F_i(\mathbf{w}^{t})$ using \\ 
 \hspace*{1em} SGD, as in (\ref{eq4}), and sends back $\Delta_i^t$ to the server
 
    \STATE  \hspace*{1em} $\mathbf{w}^{t+1}$, $\bar{ \mathcal{S}_t} \leftarrow$ \textsc{Optimal Aggregation}

\STATE \hspace*{1em} Server updates the probability $p_i^{t} \, i = 1, \cdots, |\mathcal{K}|$ by (\ref{eq7}) \\\hspace*{1em} and (\ref{eq8}) for next round's usage  

\STATE \textbf{return} $\mathbf{w}^{T}$  \\

   \renewcommand{\algorithmicrequire}{\textbf{Procedure} \	\textsc{Optimal Aggregation}}
 \renewcommand{\algorithmicensure}{\textbf{Input:} $\mathcal{S}_t$, $\Delta_i^t$, $v$, temp = $\{\}$}
 \REQUIRE 
 \ENSURE
  \STATE Direct to Algorithm \ref{algorithm1}
  \STATE \textbf{return} $\mathbf{w}^{t+1}$, $\bar{ \mathcal{S}_t}$ 
  
 \end{algorithmic} 
 \end{algorithm}
 
After getting the probability change for the labeled nodes (i.e., $i \in \mathcal{S}_t - \bar{ \mathcal{S}_t}$), we equally increase the probability for all the rest nodes $i \in \mathcal{K} - (\mathcal{S}_t - \bar{ \mathcal{S}_t})$, as shown in (\ref{eq8}).
\begin{align*}
\label{eq8} 
 p_i^{t+1} =
    \begin{cases}
       p_i^t  - \Delta p_i^t  & i \in \mathcal{S}_t - \bar{ \mathcal{S}_t} \\          
 p_i^t  + \frac{\sum_{i \in \mathcal{S}_t - \bar{ \mathcal{S}_t}} \Delta p_i^t  }{|\mathcal{K} - (\mathcal{S}_t - \bar{ \mathcal{S}_t})|}  &  i \in \mathcal{K} - (\mathcal{S}_t - \bar{ \mathcal{S}_t} )
    \end{cases}  
 ,    \tag{8}
\end{align*}
where $ p_i^{t+1}, i\in \mathcal{K}$ are used for the ($t+1$)-th round. 

We summarize the proposed FL design with probabilistic node selection and optimal aggregation in Algorithm \ref{algorithm2}.
Particularly, in each commutation round $t$, after the server receives the local update from participating nodes $i \in \mathcal{S}_t$, the server identifies the nodes that are labeled by the procedure \textsc{check expectation} (i.e., $\mathcal{S}_t - \bar{ \mathcal{S}_t}$) and the remaining nodes for aggregation $\mathcal{S}_t^* $, which are used to regulate the probability for subsequent rounds (line 7) and aggregate the global model for this round (line 6).
\subsection{Convergence Rate of \texttt{FedPNS}}
\label{IV-D}
To facilitate theoretical analysis, we introduce the auxiliary parameter $\mathbf{v}^t$, which
is optimized w.r.t. the global loss function $F( \mathbf{v})$ in the centralized setting. $\mathbf{v}^t$ is a virtual sequence since $F( \mathbf{v})$ is only observable when all data samples are available at a central place. 
We use $\tilde{\mathbf{w}}^{t}$ to denote model weight with full node participation, i.e., $\tilde{\mathbf{w}}^{t} = \sum_{i=1}^{|\mathcal{K}|} \frac{1}{|\mathcal{K}|} \mathbf{w}_i^t$.
We define that $\mathbf{v}^t$ is ``synchronized'' with $\tilde{\mathbf{w}}^{t}$ at the beginning of each global round, i.e., 
at the beginning of the $t$-th global round, the initial value of $\mathbf{v}^t$ is set as $\mathbf{v}^{t-1} = \tilde{\mathbf{w}}^{t-1}$. At the end of the $t$-th global round, the update rule of the centralized SGD is as follows
 \begin{align*}
\label{eq9}
 \mathbf{v}^t =  \mathbf{v}^{t-1} - \eta (-  \sum_{j=1}^C q (y=j) \mathbb{E}_{\mathbf{x}| y=j} \left[ {\rm log}l_j (\mathbf{v}^{t-1},\mathbf {x}, y) \right]), \tag{9}
\end{align*}
where $q (y=j)$ is the population distribution over class $j$. 

We first quantify the weight divergence $\mathbb{E}_{\mathcal{S}_t} \Vert \mathbf{w}^{t} - \mathbf{v}^{t} \Vert$ between $\mathbf{w}^{t}$ and $\mathbf{v}^{t}$, for any global round $t, t = 1, \cdots, T$. Then, by combing the result in \cite{wang2019adaptive}, we obtain the convergence rate of \texttt{FedPNS}. 

\begin{customthm}{1}
\label{thm2}
Consider $\mathcal{K}$ local nodes with equal data size and the data samples on node $i\in \mathcal{K}$ follow the data distribution $q^{(i)} $. Let assumptions \ref{ass1} and \ref{ass3} hold. Assume a fixed number of local updates $\tau$  exists between two consecutive global rounds. Then, the weight divergence in \texttt{FedPNS} after the $(t-1)$-th synchronization satisfies
\begin{align*}
\label{eq10}
  \mathbb{E}_{\mathcal{S}_t} \Vert \mathbf{w}^{t} - \mathbf{v}^{t} \Vert  \leq  \eta \sum_{i=1}^{|\mathcal{K}|}  (p_i\gamma_i + \frac{1}{|\mathcal{K}|} q_{dif}^{(i)}  ( \sum_{k=1}^{\tau-1} a^{k} g_{max}(\mathbf{v}^{t\tau-1-k}) )), \tag{10}
\end{align*}
where $g_{max}(\mathbf{v}) = max_{j=1}^C \Vert \nabla \mathbb{E}_{\mathbf{x}|  y=j} \left[ {\rm log}l_j (\mathbf{v}, \mathbf {x}, y) \right] \Vert $, $ a = 1 + \eta L$, and $q_{dif}^{(i)} = \sum_{j=1}^C  \Vert (q^{(i)} (y=j) - q (y=j)) \Vert$.
\end{customthm}

\begin{rem}
\label{rem1}
\rm
The weight divergence between $\mathbf{w}^{t}$ and $\mathbf{v}^{t}$ mainly comes from two parts, the bound of the norm of local gradient from each participating node, i.e., $\sum_{i=1}^{|\mathcal{K}|} p_i\gamma_i$, and the weight divergence introduced by the difference between the data distribution on node and population distribution, i.e., $q_{dif}^{(i)}$. \texttt{FedPNS} preferentially selects nodes with a smaller bounded gradient, which results in a smaller weight divergence, compared with node selection with equal probability in \texttt{FedAvg}, i.e., $\sum_{i=1}^{|\mathcal{K}|} p_i \gamma_i \leq \sum_{i=1}^{|\mathcal{K}|} \frac{1}{|\mathcal{K}|} \gamma_i$.  
\end{rem}

\begin{customthm}{2}
\label{thm3}
When $\eta \leq \frac{1}{L}$, compared with \texttt{FedAvg}, \texttt{FedPNS} with a smaller weight divergence achieves tighter upper bound after $T$ global rounds, i.e., $ F( \mathbf{w}^T) - F( \mathbf{w}^*) $, where $F( \mathbf{w}^*)$ denotes the optimal model parameter that minimizes $F( \mathbf{w})$.
\end{customthm}

\noindent \textit{Proof.} Theorem \ref{thm3} is proven by combing the weight divergence $\mathbb{E}_{\mathcal{S}_t} \Vert \mathbf{w}^{t} - \mathbf{v}^{t} \Vert$ in Theorem \ref{thm2} with the result in \cite[Theorem 2]{wang2019adaptive}. From Theorem \ref{thm2}, it is straightforward to see that the weight divergence $\mathbb{E}_{\mathcal{S}_t} \Vert \mathbf{w}^{t} - \mathbf{v}^{t} \Vert$ in \texttt{FedPNS} is smaller than that in \texttt{FedAvg}. From \cite[Theorem 2]{wang2019adaptive},  we have $ F( \mathbf{w}^T) - F( \mathbf{w}^*) \propto \mathbb{E}_{\mathcal{S}_t} \Vert \mathbf{w}^{t} - \mathbf{v}^{t} \Vert$, i.e., a smaller weight divergence in each global round $t, t = 1, \cdots, T$ results in a smaller gap between the global loss after $T$ global round and the global loss with optimal model, $ F( \mathbf{w}^T) - F( \mathbf{w}^*)$, which completes the proof.

\subsection{Complexity Analysis}
\label{IV-E}
{We consider the model in \textit{float} format (i.e., 32 bits for each parameter) and the operations in algorithms are float point operations. For simplicity, we consider a general $n_{layer}$ layers fully connected neural network (FCNN) with the same number of parameters $n $, in each layer (i.e., the total parameters of model update/gradient is $n \cdot n_{layer} $ and $n\gg n_{layer}$ holds typically). The output of the $k$-th layer in forward propagation (FP) is represented as $a^{(k)} = g(z^{(k)}), z^{(k)} = \mathbf{w}^{(k)}  a^{(k-1)} $, where $g(\cdot)$ is the activation function which is evaluated elementwise, $\mathbf{w}^{(k)}$ is the model parameter in the $k$-th layer, and the bias component in FCNN is omitted for simplicity. We assume the number of features for the input layer is $n$. The computation in each layer is viewed as a matrix-vector multiplication, and an activation function, thus the complexity for multiplications in FP and for activation function applied in FP are
$\sum_{k=1}^{n_{layer}} n^{2} = n_{layer} \cdot n^2$ and $\sum_{k=1}^{n_{layer}} n = n_{layer} \cdot n$, respectively. Therefore, the complexity for forward propagation of FCNN (also for \textsc{check loss}, line 24) is $\mathcal{O}(n^2)$ since $n\gg n_{layer} $. Given the total $n \cdot n_{layer} $ parameters of local gradient, the complexity for arithmetic addition and arithmetic multiplication are $\mathcal{O}(n)$ and $\mathcal{O}(n^2)$, respectively. 

With regards to the complexity of the proposed Algorithm 1, we consider the procedures \textsc{Check Expectation} and \textsc{Check loss}. In particular, generating a global gradients/model needs $|\mathcal{S}_t|$ additions (line 20, line 23) and $|\mathcal{S}_t|$ multiplications are needed for calculating the expectation values (i.e., temp, line 21). As such, the complexity for \textsc{Check Expectation} is  $\mathcal{O}(n^2)$} since the number of local updates $|\mathcal{S}_t|$ and the number of iterations $|\mathcal{S}_t|-v$ (line 4) are much smaller than $n$. The complexity for Algorithm 1 is $\mathcal{O} (n^2 + n^2 + n) = \Theta (2n^2 + n)$. In Algorithm 2, the complexity for adjusting the probability (line 7) is $\mathcal{O} (|\mathcal{K}|)$, which is marginal compared with the complexity of \textsc{Optimal Aggregation} (line 6). Therefore, the complexity for Algorithm 2 is $\mathcal{O} (n^2 + n^2 + n) = \Theta (2n^2 + n)$. Compared with local training including FP and back propagation (BP) (the complexity for BP is $\mathcal{O} (n^3)$) at node side, the overhead of the proposed algorithms that are conducted at server side is marginal and can be ignored. 

\section{Evaluation and analysis}
\label{V}
We now present empirical results for the proposed probabilistic node selection strategy. We implement  \texttt{FedPNS} on different tasks, models, datasets, and compare with commonly used benchmark \texttt{FedAvg}. We first demonstrate the effectiveness of the proposed \texttt{Optimal Aggregation} in enlarging the expected decrement of FL global loss and in identifying the potential adverse local updates (Section \ref{V-A}). Then, the superiority of the proposed \texttt{FedPNS} in presence of different data heterogeneity is illustrated in Section \ref{V-B}. The choice $\alpha$ and $\beta$ for adjusting node probability in \texttt{FedPNS} are discussed in Section \ref{V-C}. In Section \ref{V-D}, by tracking the norm of gradient on different nodes, the Assumption \ref{ass3} is empirically justified. In addition, we also compare the proposed \texttt{FedPNS} with an existing work that uses the norm of gradient $\Vert \nabla F_i(\mathbf{w}^{t}) \Vert$ \cite{9337227} for node selection. All code, data, and experiments are publicly available as an open-source GitHub repository at: \href{https://github.com/HongdaWu1226/FedPNS}{github.com/HongdaWu1226/FedPNS}.

We briefly describe our adopted datasets, learning model, and experiment setting as follows.

\textbf{Synthetic data.} To better characterize the data heterogeneity and study its impact on model convergence, we generate synthetic data by following the similar setup as in \cite{li2018federated, shamir2014communication}. In particular, the data samples $\{\mathbf{x}, y\}$ on local node $i$ are generated according to the model $y= argmax({\rm softmax}(\mathbf{wx} +b )), \mathbf{x} \in \mathbb{R}^{60}, \mathbf{w} \in \mathbb{R}^{10 \times 60},  b \in \mathbb{R}^{10}$. We set $\mathbf{w}, b \backsim \mathcal{N}(0,1)$. For the data on i.i.d. nodes, $\mathbf{x}$ follows the same distribution $\mathcal{N}(0,\Sigma)$, where $\Sigma$ is diagonal with $\Sigma_{r,r} = r^{-1.2}$. For the data samples on non-i.i.d. node $i$, $\mathbf{x} \backsim \mathcal{N}(o_i,\Sigma)$, each element in the mean vector $o_i$ is drawn from $\mathcal{N}(B_i, 1), B_i \backsim N(0, \varrho)$. As such, a big value of $\varrho$ denotes a more heterogeneous data scenario. The training set and testing set are randomly split with $80\%-20\%$ proportion on each node. A Multinomial Logistic Regression (MLR) model is applied to the synthetic data.

\textbf{Real data.} We explore different learning objectives on different real datasets, which are considered in prior works \cite{McMahan2017CommunicationEfficientLO, li2018federated}. In Section \ref{V-B}, we start with a convex classification problem with MNIST \cite{lecun-mnisthandwrittendigit-2010} using MLR model. Then, for the non-convex setting, we consider two CNN models for MNIST and CIFAR-10\cite{cifar}, which are referred as CNN-M\footnote{The CNN-M model has 7 layers with the following structure: $\rm 5 \times 5 \times 10$ Convolutional $\rightarrow 2 \times 2$ MaxPool $\rightarrow $ $\rm 5 \times 5 \times 20$ Convolutional $\rightarrow 2 \times 2$ MaxPool $\rightarrow 320\times50 $ Fully connected $\rightarrow 50\times10 $ Fully connected $\rightarrow$ Softmax. The second convolutional layer is with 50\% dropout. All Convolutional and Fully connected layers are mapped by ReLu activation.} and CNN-C\footnote{The CNN-C model has 8 layers as structured follows: $\rm 5 \times 5 \times 6$ Convolutional $\rightarrow 2 \times 2$ MaxPool $\rightarrow $ $\rm 5 \times 5 \times 16$ Convolutional $\rightarrow 2 \times 2$ MaxPool $\rightarrow 400\times120 $ Fully connected $\rightarrow 120\times84 $ Fully connected $\rightarrow 84\times10 $ Fully connected $\rightarrow$ Softmax. ReLu activation is applied to all layers.} hereinafter.

Through the experimental result, unless otherwise specified, we evaluate the accuracy of the trained models using the testing set from each dataset. The fraction for selecting nodes is set to be $c=0.2, |\mathcal{S}_t|= c\mathcal{|K|} = 10$, $D_i=200$, $ B=20$, $E=1$, $T=200$, $\eta=0.01$, decay rate $=0.995$, $v=0.7$, $ \bar B=128$. For real datasets, the overall data heterogeneity is measured by $\sigma$ and the skewness of dataset on non-i.i.d. nodes is represented by $\rho$. For example, $\sigma = 0.2, \rho = 2$ means that $\sigma \mathcal{|K|} = 10$ nodes are equipped with i.i.d. dataset, where non-i.i.d. dataset lay on the rest   $(1-\sigma) \mathcal{|K|} = 40$ nodes, and the data samples on which are  evenly belong to 2 labels. As such, a small $\sigma, \rho$ indicates a higher data heterogeneity.

\begin{figure}[t]
\centering 
{\includegraphics[scale=0.43]{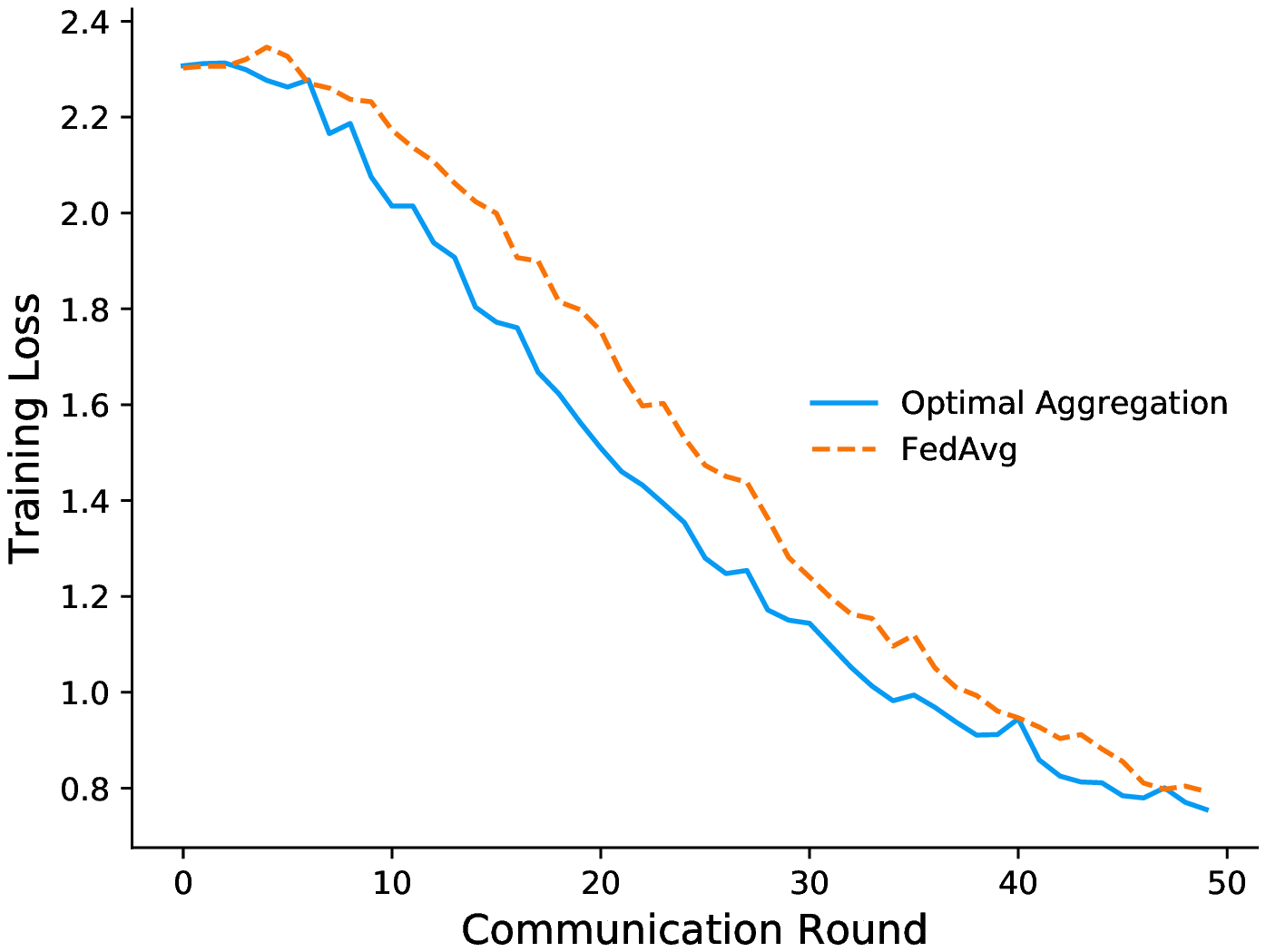}}
{\includegraphics[width=8cm,height= 5cm]{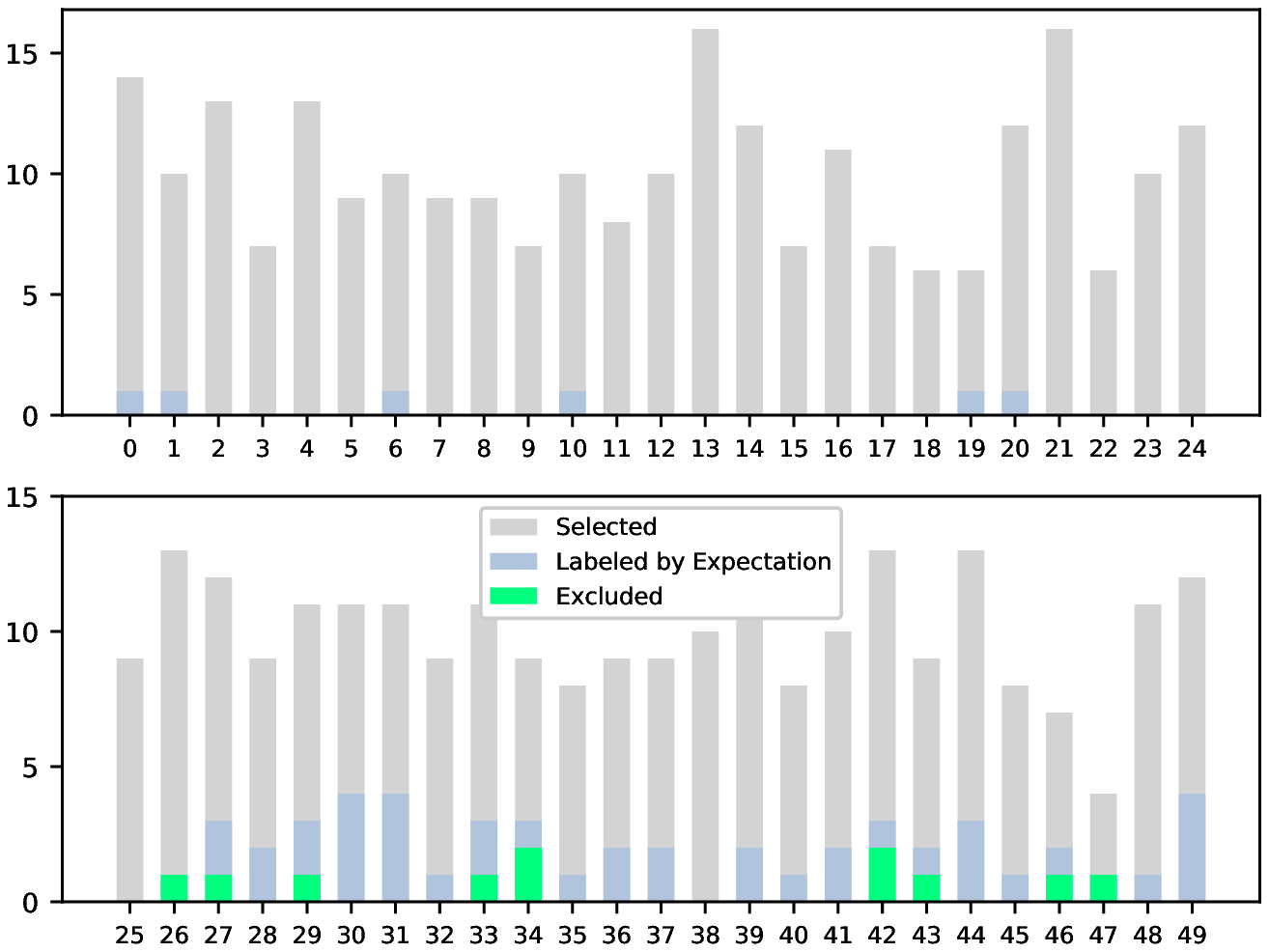}}
\caption{Performance of the proposed \texttt{Optimal Aggregation}. (1) Upper: The training loss on the MNIST dataset when different aggregation strategies are adopted. \texttt{Optimal Aggregation} and \texttt{FedAvg} aggregate local updates over $\mathcal{S}^*_t$ and $\mathcal{S}_t$, respectively. (2) Bottom: We use a triple to observe the result of \texttt{Optimal Aggregation}, which includes
 the accumulated times that each node is selected, labeled by \textsc{Check expectation}, and excluded eventually by \textsc{Check loss} during FL model training. The upper and bottom row refer to the results for i.i.d. nodes and non-i.i.d. nodes, respectively.}
  \label{fig1}
\end{figure}
\subsection{Performance of \texttt{Optimal Aggregation}}
\label{V-A}
In this part, we conduct an experiment to illustrate the performance of the proposed \texttt{Optimal Aggregation} algorithm. Particularly, we adopt CNN-M model on MNIST dataset where the data heterogeneity is set to be  $\sigma = 0.5, \rho = 1$. In each global round, we randomly select $|\mathcal{S}_t| = 10$ nodes while guaranteeing the participating nodes include half i.i.d. nodes and half non-i.i.d. nodes. To avoid the randomness of node selection, the participating nodes in each round are kept as the same for \texttt{FedAvg}\cite{li2018federated} and the proposed \texttt{Optimal Aggregation} algorithm.

\begin{figure*}[t]
\centerline{\includegraphics[scale=0.33]{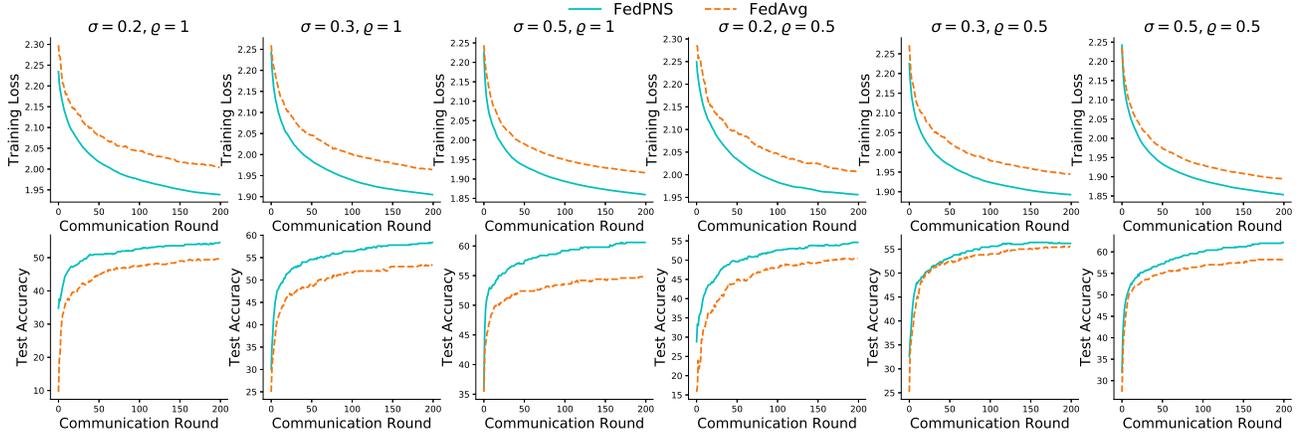}}
\caption{Effect of data heterogeneity on convergence. (1) Top row: we show the training loss on synthetic dataset whose data heterogeneity decreases from left to right (with a fixed $\sigma$ or $\varrho$). (1) Bottom row: we show the corresponding test accuracy. } 
\label{fig2}
\end{figure*}
\begin{figure*}[h]
\centerline{\includegraphics[scale=0.33]{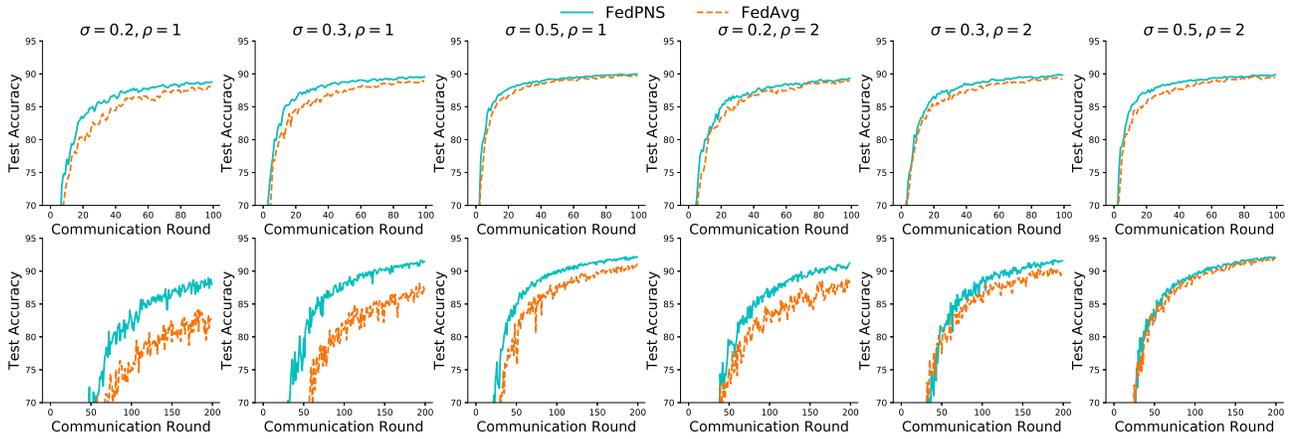}}
\caption{Test accuracy over communcation rounds of \texttt{FedPNS} and \texttt{FedAvg} with different data heterogeneity. Upper and lower subplots correspond to training performance when the MLR model and CNN-M model are adopted for MNIST, respectively. A smaller $\alpha, \rho$ indicates a higher data heterogeneity.}
\label{fig3}
\end{figure*}

As shown in the upper part of Fig. \ref{fig1}, the proposed \texttt{Optimal Aggregation} algorithm can achieve lower training loss than \texttt{FedAvg}. When the global model is not robust in the several initial rounds, the local updates are more diverse due to the data heterogeneity, thus excluding adverse local updates is more effective. We count the accumulated times that each node is selected, labeled by the procedure \textsc{Check expectation} (line 7 in Algorithm \ref{algorithm1}), and finally excluded by the procedure \textsc{Check loss} (line 14 in Algorithm \ref{algorithm1}). As we can see from the bottom part of Fig. \ref{fig1}, i) the i.i.d nodes (i.e., with index $``0", \cdots, ``24"$) are never been excluded, yet some of the non-i.i.d nodes (e.g.,$``26",``27", ``34"$, etc.) have been excluded for many times. ii) Almost all non-i.i.d. nodes are labeled at least one time, which illustrates the effectiveness of \texttt{Optimal Aggregation} in identifying the nodes with the skewed dataset.

\subsection{Data Heterogeneity}
\label{V-B}
In this section, we use different combinations of $\sigma$ and  $\rho$ to investigate the performance of the proposed \texttt{FedPNS} scheme in presence of different data heterogeneity. Through all experiments, $\alpha$ and $ \beta$ are chosen to be 2 and 0.7 respectively. The choice $\alpha$ and $\beta$ are discussed in Section \ref{V-C}.

\subsubsection{MLR Model with Synthetic Data}
We follow the description in Section \ref{V-A} to generate synthetic data samples. 
The ratio of i.i.d. nodes is set to be $\sigma=0.2, 0.3,$ and $0.5$ with $\varrho = 0.5$ and $1$. For each node $i$, the number of data samples $D_i =1000$ and the number of epochs for local training is $E=20$. In Fig. \ref{fig2}, we study how data heterogeneity affects model convergence using MLR model and synthetic dataset. As we can see from Fig. \ref{fig2}, as the data heterogeneity increases, i.e., $\sigma=0.5,0.3$ and $0.2$ with fixed $\varrho=1$ or $0.5$, \texttt{FedAvg} slows to converge (i.e., higher training loss) with a decreasing test accuracy in the meantime. \texttt{FedPNS} achieves a lower training loss and higher test accuracy, compared with \texttt{FedAvg} in all data setting.

\begin{figure*}[t]
\centerline{\includegraphics[scale=0.32]{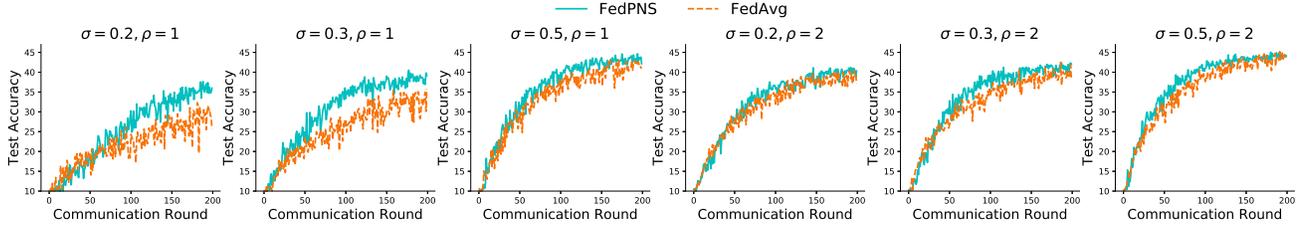}}
\caption{Test accuracy over communcation rounds of \texttt{FedPNS} and \texttt{FedAvg} with different data heterogeneity. CNN-C model is adopted for CIFAR-10.}
\label{fig4}
\end{figure*}

\subsubsection{MLR, CNN-M Model for MNIST}
As we can tell from Fig. \ref{fig3}, \texttt{FedPNS} converges faster and achieves a higher test accuracy, compared with \texttt{FedAvg} for both MLR and CNN model regardless of different data heterogeneity. \texttt{FedPNS} achieves better improvement when the CNN model is adopted, compared with the scenario when the MLR model is utilized, which attributes to the limited learning capability of MLR. In addition, it is observable that as the data becomes more heterogeneous, the performance enhancement is enlarged (i.e., $\alpha$ decreases from 0.5 to 0.2 for a given $\beta$, or $\beta$ changes from 2 to 1 for a given $\alpha$).
When the number of i.i.d. nodes is limited and the non-i.i.d nodes are equipped with highly skewed dataset (e.g., $\sigma=0.2, \rho = 1$ and $\sigma=0.3, \rho = 1$), \texttt{FedPNS} gains remarkable performance improvement, which verifies the effectiveness of \texttt{FedPNS} in identifying and selecting the nodes that contribute global model better. For the scenario with the lowest data heterogeneity (i.e., $\sigma=0.5, \rho=2$), the performance gap between \texttt{FedPNS} and \texttt{FedAvg} is not obvious. This is because the impact of the non-i.i.d. nodes on the convergence is reduced when a large number of i.i.d. nodes can be selected.

\subsubsection{CNN-C Model for CIFAR-10}
For the more complex three channel image classification task, the number of local epoch is set to be $E=5$. As we can see from Fig. \ref{fig4}, compared with \texttt{FedAvg}, \texttt{FedPNS} converges faster and leads to a higher test accuracy, especially for the high data heterogeneity scenario (i.e., $\sigma=0.2$ and $0.3, \rho=1$). The performance improvement of \texttt{FedPNS} is not obvious when $\sigma=0.2, \rho=2$, this is because the small number of i.i.d. nodes with less heterogeneous data samples on non-i.i.d. nodes makes \texttt{FedPNS} hard to distinguish the node contribution.

\subsection{Choosing $\alpha$ and $\beta$}
\label{V-C}
The choice of $\alpha$ and $\beta$ gives an impact on \texttt{FedPNS}. As discussed in Section \ref{IV-C}, a large value of $\alpha$ can help increase the model convergence rate by aggressively adjusting the node probability. On the other hand, a large value of $\alpha$ also makes node selection sensitive to the identification mistake, which may negatively impact the convergence. A similar effect is achieved by $\beta$, which keeps the rate of probability change in a range $[\beta, 1]$.
\begin{figure}[h]
\centering 
{\includegraphics[scale=0.32]{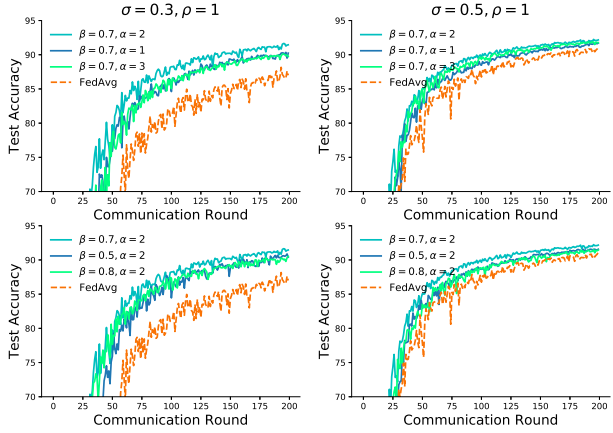}}
\caption{Effect of adopting different $\alpha$ and $\beta$. We heuristically choosing $\alpha \in \mathbb{Z}^+, \beta\in [0,1]$ in ascending order. The top row and bottom row correspond to the performance with varied $\alpha$ and $\beta$, respectively. CNN-M on MNIST is adopted.}
  \label{fig5}
\end{figure}
We studied the effect of different $\alpha$ and $\beta$ via heuristically choosing $\alpha \in \mathbb{Z}^+, \beta\in [0,1]$ in ascending order. From the top row of Fig. \ref{fig5}, for a fixed $\beta =0.7$, increasing $\alpha$ from 1 to 2 boosts performance. However, keep increasing $\alpha$ does not consistently embrace performance gain, this is because \texttt{FedPNS} becomes more sensitive to identification mistakes, which may prevent i.i.d nodes from being selected in the subsequent rounds. Similarly, from the bottom plot of Fig. \ref{fig5}, for a fixed $\alpha =2$, increasing $\beta$ from 0.5 to 0.7 promotes model performance. However, further increasing $\beta$ to 0.8 leads to a degraded performance. Empirically, we find $\alpha=2, \beta=0.7$ that balances the tradeoff and leads to the best performance.

\subsection{Other Comparison}
\label{V-D}
In this section, we take one experimental case as an example to demonstrate the bounded norm of local gradient $\Vert \nabla F_i(\mathbf{w}^t) \Vert$, which is related to the data distribution on each node. Besides, we compare the proposed \texttt{FedPNS} with another node selection scheme \texttt{BN2} \cite{9337227}, which chooses the nodes with higher $\Vert \nabla F_i(\mathbf{w}^t) \Vert$ for aggregation. Specifically, in each global round, \texttt{BN2} first randomly selects $|\mathcal{M}|$ nodes for local training. After that, the participating nodes send their gradient norm $\Vert \nabla F_i(\mathbf{w}^t) \Vert, i \in \mathcal{M}$ to the server. The server chooses the first $|\mathcal{S}_t|$ local updates for model aggregation by sorting $\Vert \nabla F_i(\mathbf{w}^t) \Vert, i \in \mathcal{M}$ in descending order.

\begin{figure}[h]
\centering 
{\includegraphics[scale=0.43]{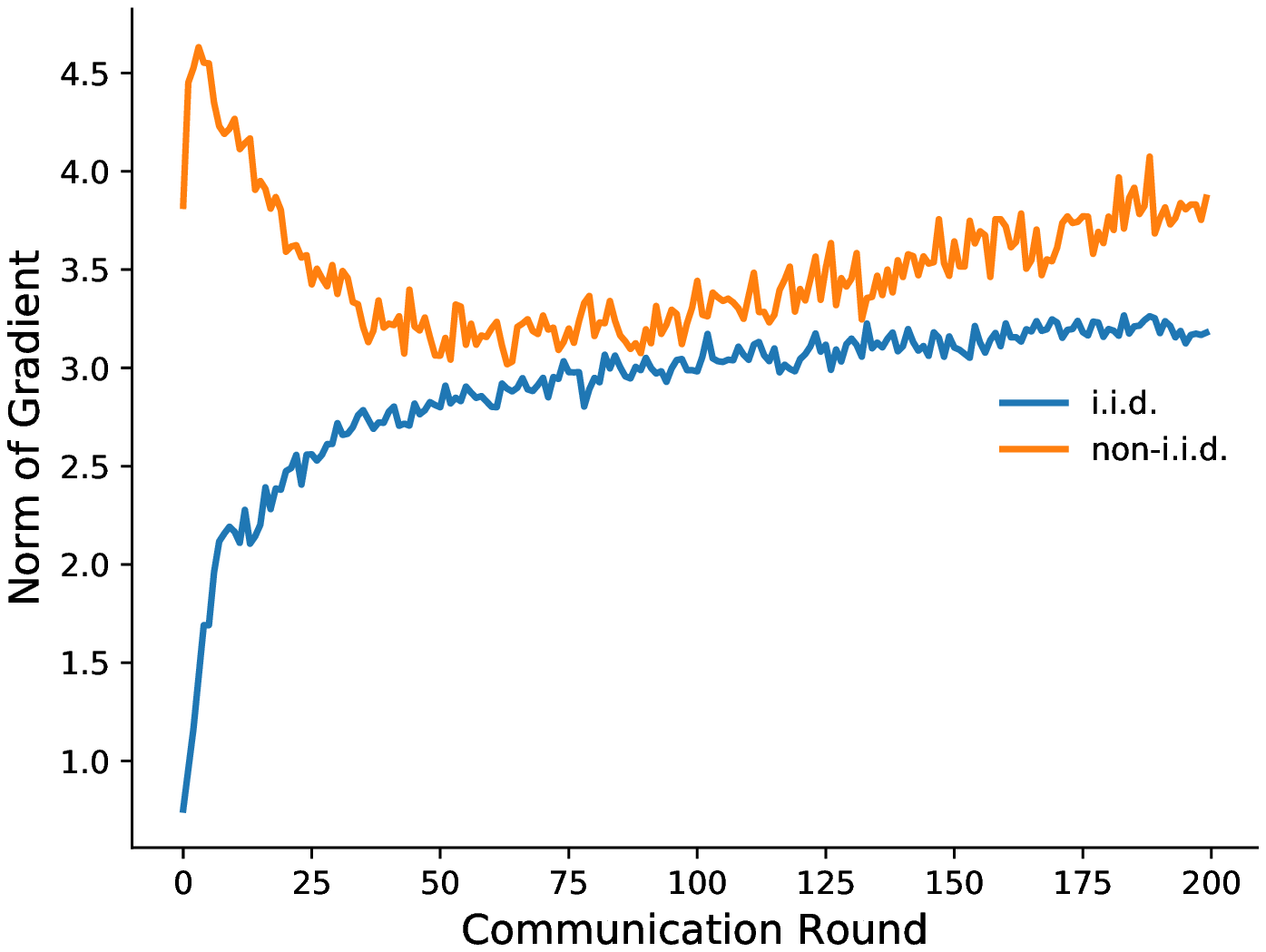}}
{\includegraphics[scale=0.43]{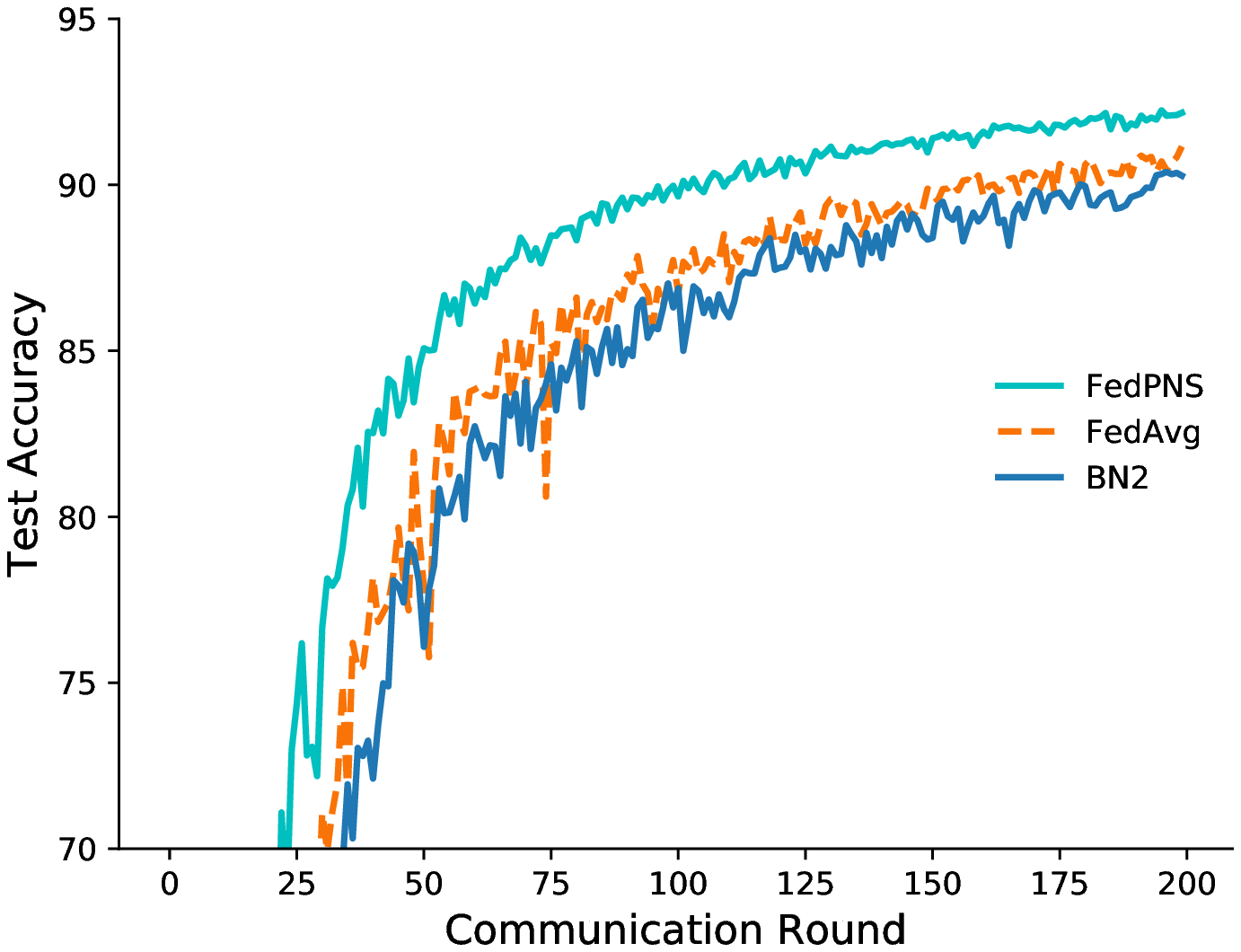}}
\caption{Node selection design with different importance indicator. \texttt{FedPNS} chooses nodes by measuring the data distribution on local nodes, while \texttt{BN2} selects nodes according to the norm of gradient. (1) Top plot: we track the averaged gradient norm of node $i\in \mathcal{M}$ with different data distribution, where each node is selected from $\mathcal{K}$ randomly. (2) Bottom plot: we compare the test accuracy for different node selection designs. CNN-M on MNIST is adopted with $\sigma=0.5, \rho=1$.}
  \label{fig6}
\end{figure}

In this experiment, $|\mathcal{M}|$ is set to be 20. We track the norm of gradient for each participating node $i \in \mathcal{M}$ statistically in each global round. As we can see from Fig. \ref{fig6}, the averaged gradient norm from i.i.d. nodes is smaller than that from non-i.i.d. nodes. This is because the data distribution on i.i.d. nodes is more similar to population distribution that is defined over all nodes. As such, preferentially scheduling the nodes with higher norm of gradient would slow the convergence, as shown in the bottom of Fig. \ref{fig6}.

\section{Conclusion}
\label{VI}
In this paper, we have presented our design of \texttt{FedPNS} algorithm, a probabilistic node selection strategy that can preferentially select nodes to boost model convergence of FL with non-i.i.d. datasets. \texttt{FedPNS} adjusts the probability for each node to be selected in each round based on the result of the proposed \texttt{Optimal Aggregation} algorithm, which is able to find out the optimal subset of local updates from participating nodes and excludes the adverse local updates for a better model aggregation, by measuring the relationship between the local gradient and the global gradient from participating nodes. The convergence rate improvement of the \texttt{FedPNS} design over \texttt{FedAvg} is analyzed theoretically. Finally, experimental results on different tasks, models, and datasets have shown that FL training with \texttt{FedPNS}
accelerates model convergences and leads to higher test accuracy, as compared to \texttt{FedAvg}.

\appendices
\section*{appendix}
\subsection{Proof of Lemma \ref{thm1}}
\label{apex1}
From the $L$-smooth of $F(\mathbf{w})$ and applying Taylor expansion, we have
\begin{align*}
\label{a1}
F(\mathbf{w}^{t+1}) \leq  F(\mathbf{w}^{t}) +  \langle \nabla F( \mathbf{w}^{t}), \mathbf{w}^{t+1} - \mathbf{w}^{t} \rangle + \frac{L}{2} \Vert\mathbf{w}^{t+1} - \mathbf{w}^{t} \Vert ^2. \tag{A1}
\end{align*}

$\bullet$ Bounding $\Vert\mathbf{w}^{t+1} - \mathbf{w}^t \Vert^2$:
By the definition of the global aggregation in (\ref{eq5}) and local update calculated by (\ref{eq4}), we have
\begin{align*}
\label{a2}
 \Vert \mathbf{w}^{t+1} - \mathbf{w}^{t}\Vert^2 
 & =  (\mathbb{E}_{i \backsim \mathcal{S}_t} \left[  \Vert \mathbf{w}^{t+1} - \mathbf{w}^{t}\Vert \right])^2 \\
 & = \eta^2  (\mathbb{E}_{i \backsim \mathcal{S}_t} \left[ \Vert  \nabla F_i( \mathbf{w}^{t}) \Vert  \right])^2\\
  &\stackrel{1}\leq  \eta^2  \mathbb{E}_{i \backsim \mathcal{S}_t} \left[ \Vert  \nabla F_i( \mathbf{w}^{t}) \Vert^2 \right] \\
  &\leq  \eta^2  \Vert \nabla F( \mathbf{w}^{t}) \Vert^2 \delta^2,
 \tag{A2}
\end{align*}
where inequality 1 holds because of Cauchy-Schwarz inequality and the last inequality is due to the bounded dissimilarity assumption.

$\bullet$ Bounding $\langle \nabla F( \mathbf{w}^t),  \mathbf{w}^{t+1} - \mathbf{w}^t \rangle$: 
Again, by the definition of the global aggregation for $\mathbf{w}^{t+1}$ and SGD optimization, we have
\begin{align*}
\label{a3}
 \langle \nabla F( \mathbf{w}^{t})),  \mathbf{w}^{t+1} - \mathbf{w}^{t}) \rangle =  -\eta \mathbb{E}_{i \backsim \mathcal{S}_t}\left[     \langle \nabla F(\mathbf{w}^{t}),  \nabla F_i(\mathbf{w}^{t})  \rangle \right].  \tag{A3}
\end{align*}

Plugging (\ref{a2}) and (\ref{a3}) into (\ref{a1}), we obtain
\begin{align*}
\label{a4}
F(\mathbf{w}^{t+1}) -  F(\mathbf{w}^{t}) & \leq -\eta \mathbb{E}_{i \backsim \mathcal{S}_t}\left[     \langle \nabla F(\mathbf{w}^{t}),  \nabla F_i(\mathbf{w}^{t})  \rangle \right] \\ & +  \frac{L\eta^2}{2}  \Vert \nabla F( \mathbf{w}^{t}) \Vert^2 \delta^2.
 \tag{A4}
\end{align*}

\subsection{Proof of Theorem \ref{thm2}}
At any global round $t$, the weight divergence between the model $\mathbf{w}^{t}$ with partial node participation and centralized model $\mathbf{v}^{t}$ is bounded as follows
\begin{align*}
\label{C1}
\mathbb{E}_{\mathcal{S}_t} \Vert \mathbf{w}^{t} - \mathbf{v}^{t} \Vert & = \mathbb{E}_{\mathcal{S}_t} \Vert \mathbf{w}^{t} - \tilde{\mathbf{w}}^{t} + \tilde{\mathbf{w}}^{t} -  \mathbf{v}^{t} \Vert \\
& \leq \mathbb{E}_{\mathcal{S}_t} \Vert \mathbf{w}^{t} - \tilde{\mathbf{w}}^{t} \Vert + \Vert \tilde{\mathbf{w}}^{t} - \mathbf{v}^{t} \Vert. \tag{B1}
\end{align*}

We will separately bound the last two terms on the right-hand side of the above inequality.

$\bullet$ Bounding $\Vert \tilde{\mathbf{w}}^{t} - \mathbf{v}^{t} \Vert$: In this part, to facilitate analysis, we introduce the index of local update, e.g., the models $\tilde{\mathbf{w}}^{t}$ and $\mathbf{v}^{t}$ are represented by $\tilde{\mathbf{w}}^{t\tau}$ and $\mathbf{v}^{t\tau}$ since $\tau$ times of local SGD are applied in each global round.  

Based on the definition of $\tilde{\mathbf{w}}^{t}$ and $\mathbf{v}^{t}$, we have
\begin{align*}
\label{C2}
&  \Vert \tilde{\mathbf{w}}^{t} - \mathbf{v}^{t} \Vert  =  \Vert \tilde{\mathbf{w}}^{t\tau} - \mathbf{v}^{t\tau} \Vert = \Vert \sum_{i=1}^{|\mathcal{K}|} \frac{D_i}{\sum_{i=1}^{|\mathcal{K}|}D_i}  \mathbf{w}_i^{t\tau} - \mathbf{v}^{t\tau} \Vert \\
\stackrel{1}= &  \Vert \sum_{i=1}^{|\mathcal{K}|} \frac{1}{|\mathcal{K}|}  (\mathbf{w}_i^{t\tau-1} - \eta \nabla F_i(\mathbf{w}_i^{t\tau-1}) ) - \mathbf{v}^{t\tau-1} +  \eta \nabla F(\mathbf{v}^{t\tau-1})\Vert\\
\stackrel{2}\leq &  \Vert \sum_{i=1}^{|\mathcal{K}|} \frac{1}{|\mathcal{K}|} \mathbf{w}_i^{t\tau-1} -  \mathbf{v}^{t\tau-1}\Vert + \eta \Vert \sum_{i=1}^{|\mathcal{K}|} \frac{1}{|\mathcal{K}|} \sum_{j=1}^C q^{(i)} (y=j) \\ 
& ( \nabla \mathbb{E}_{\mathbf{x}| y=j} \left[ {\rm log}l_j (\mathbf{w}_i^{t\tau-1}, \mathbf {x}, y) \right] - \nabla \mathbb{E}_{\mathbf{x}| y=j} \left[ {\rm log}l_j (\mathbf{v}^{t\tau-1}, \mathbf {x}, y) \right]) \Vert  \\
\stackrel{3} = &  \Vert \sum_{i=1}^{|\mathcal{K}|} \frac{1}{|\mathcal{K}|} \mathbf{w}_i^{t\tau-1} -  \mathbf{v}^{t\tau-1}\Vert + \eta \Vert \sum_{i=1}^{|\mathcal{K}|} \frac{1}{|\mathcal{K}|}  (\nabla  F_i(\mathbf{w}_i^{t\tau-1}) -\nabla  F_i(\mathbf{v}^{t\tau-1})) \Vert \\
\stackrel{4}\leq & \sum_{i=1}^{|\mathcal{K}|} \frac{1}{|\mathcal{K}|} (1 + \eta L ) \Vert \mathbf{w}_i^{t\tau-1} - \mathbf{v}^{t\tau-1} \Vert, \tag{B2}
\end{align*}
where equality 1 holds by the updating rule of SGD and by that all nodes are with equal data size. Inequality 2 holds by applying triangle inequality and by the observation that for each class, the data distribution over all nodes is the same as the distribution over the whole data samples, i.e., $j\in [C], q (y=j) = \sum_{i=1}^{|\mathcal{K}|} \frac{1}{|\mathcal{K}|} q^{(i)} (y=j) $. Equality 3 holds by (\ref{eq2}), (\ref{eq3}) and (\ref{eq9}). and inequality 4 holds by Assumption \ref{ass1} that the local loss function is $L$-smooth.

For node $i\in \mathcal{K}$, $ \Vert \mathbf{w}_i^{t\tau-1} - \mathbf{v}^{t\tau-1} \Vert$ is bounded as
\begin{align*}
\label{C3}
& \Vert \mathbf{w}_i^{t\tau-1} - \mathbf{v}^{t\tau-1} \Vert \\
 = & \Vert \mathbf{w}_i^{t\tau-2} - \eta \nabla F_i(\mathbf{w}_i^{t\tau-2})  - \mathbf{v}^{t\tau-2} +  \eta \nabla F(\mathbf{v}^{t\tau-2})\Vert\\
\leq & \Vert  \mathbf{w}_i^{t\tau-2} -  \mathbf{v}^{t\tau-2}\Vert  + \eta \Vert \sum_{j=1}^C q^{(i)} (y=j)   \nabla \mathbb{E}_{\mathbf{x}|  y=j} \left[ {\rm log}l_j (\mathbf{w}_i^{t\tau-2}, \mathbf {x}, y) \right] \\ &
    - \sum_{j=1}^C q (y=j) \nabla \mathbb{E}_{\mathbf{x}|  y=j} \left[ {\rm log}l_j (\mathbf{v}^{t\tau-2}, \mathbf {x}, y) \right] \Vert\\
  \stackrel{5}\leq & \Vert  \mathbf{w}_i^{t\tau-2} -  \mathbf{v}^{t\tau-2}\Vert + \eta \Vert \sum_{j=1}^C q^{(i)} (y=j) \\ & (\nabla \mathbb{E}_{\mathbf{x}|  y=j} \left[ {\rm log}l_j (\mathbf{w}_i^{t\tau-2}, \mathbf {x}, y) \right] 
   - \nabla \mathbb{E}_{\mathbf{x}|  y=j} \left[ {\rm log}l_j (\mathbf{v}^{t\tau-2}, \mathbf {x}, y) \right]) \Vert \\ 
   & +  \eta \Vert \sum_{j=1}^C (q^{(i)} (y=j) - q (y=j))   \nabla \mathbb{E}_{\mathbf{x}|  y=j} \left[ {\rm log}l_j (\mathbf{v}^{t\tau-2}, \mathbf {x}, y) \right] \Vert \\
     \stackrel{6}= & \Vert  \mathbf{w}_i^{t\tau-2} -  \mathbf{v}^{t\tau-2}\Vert + \eta \Vert  \nabla  F_i(\mathbf{w}_i^{t\tau-2}) -\nabla  F_i(\mathbf{v}^{t\tau-2}) \Vert \\ 
   & +  \eta \Vert \sum_{j=1}^C (q^{(i)} (y=j) - q (y=j))   \nabla \mathbb{E}_{\mathbf{x}|  y=j} \left[ {\rm log}l_j (\mathbf{v}^{t\tau-2}, \mathbf {x}, y) \right] \Vert \\
\stackrel{7}\leq & (1 + \eta L ) \Vert \mathbf{w}_i^{t\tau-2} - \mathbf{v}^{t\tau-2} \Vert  \\
 & + \eta g_{max}(\mathbf{v}^{t\tau-2})  \sum_{j=1}^C \Vert (q^{(i)} (y=j) - q (y=j)) \Vert, \tag{B3}
\end{align*}
where inequality 5 holds by introducing a term 
$\sum_{j=1}^C q^{(i)} (y=j) \nabla \mathbb{E}_{\mathbf{x}|  y=j} \left[ {\rm log}l_j (\mathbf{v}^{t\tau-2}, \mathbf {x}, y) \right]$ and applying triangle inequality. Equality 6 holds by (\ref{eq2}), (\ref{eq3}) and (\ref{eq9}). Inequality 7 holds by Assumption \ref{ass1} and by defining $g_{max}(\mathbf{v}^{t\tau-2}) = max_{j=1}^C \Vert \nabla \mathbb{E}_{\mathbf{x}|  y=j} \left[ {\rm log}l_j (\mathbf{v}^{t\tau-2}, \mathbf {x}, y) \right] \Vert $.

Based on (\ref{C3}), by mathematical induction and setting $ a = 1 + \eta L$, we have
\begingroup
\allowdisplaybreaks
\begin{align*}
\label{C4}
& \Vert \mathbf{w}_i^{t\tau-1} - \mathbf{v}^{t\tau-1} \Vert \\
\leq & \quad  a \Vert \mathbf{w}_i^{t\tau-2} - \mathbf{v}^{t\tau-2} \Vert  \\
& + \eta  \sum_{j=1}^C \Vert (q^{(i)} (y=j) - q (y=j)) \Vert g_{max}(\mathbf{v}^{t\tau-2}) \\
\leq & \quad  a^2 \Vert \mathbf{w}_i^{t\tau-3} - \mathbf{v}^{t\tau-3} \Vert  + \eta \sum_{j=1}^C \Vert (q^{(i)} (y=j) - q (y=j)) \Vert \\ & \qquad \qquad  \qquad \qquad  (g_{max}(\mathbf{v}^{t\tau-2}) + a g_{max}(\mathbf{v}^{t\tau-3}))\\
\vdots \\
\leq & \quad  a^{\tau-1} \Vert \mathbf{w}_i^{(t-1)\tau} - \mathbf{v}^{(t-1)\tau} \Vert  + \eta \sum_{j=1}^C \Vert (q^{(i)} (y=j) - q (y=j)) \Vert \\ & \qquad \qquad  \qquad \qquad  ( \sum_{k=0}^{\tau-2} a^{k} g_{max}(\mathbf{v}^{t\tau-2-k}). \tag{B4}
\end{align*}

Substituting (\ref{C4}) to (\ref{C2}), we obtain
\begin{align*}
\label{C5}
 \Vert \tilde{\mathbf{w}}^{t} - \mathbf{v}^{t} \Vert &  \leq \sum_{i=1}^{|\mathcal{K}|} \frac{1}{|\mathcal{K}|} ( a^{\tau} \Vert \mathbf{w}_i^{(t-1)\tau} - \mathbf{v}^{(t-1)\tau} \Vert \\ + \eta \sum_{j=1}^C  & \Vert (q^{(i)} (y=j) - q (y=j)) \Vert ( \sum_{k=1}^{\tau-1} a^{k} g_{max}(\mathbf{v}^{t\tau-1-k}))). \tag{B5}
\end{align*}
\endgroup

Since $\mathbf{v}^t$ is ``synchronized'' with $\tilde{\mathbf{w}}^{t}$ at the beginning of each global round, we ignore the first item of the right hand side of (\ref{C5}), which is the weight divergence accumulated from the previous round. Thus, the weight divergence $\Vert \tilde{\mathbf{w}}^{t} - \mathbf{v}^{t} \Vert$ between two consecutive global round is represented as
\begin{align*}
\label{C6}
 \Vert \tilde{\mathbf{w}}^{t} - \mathbf{v}^{t} \Vert   \leq  \eta \sum_{i=1}^{|\mathcal{K}|} \frac{1}{|\mathcal{K}|} q_{dif}^{(i)}  ( \sum_{k=1}^{\tau-1} a^{k} g_{max}(\mathbf{v}^{t\tau-1-k}), \tag{B6}
\end{align*}
where $q_{dif}^{(i)} = \sum_{j=1}^C  \Vert (q^{(i)} (y=j) - q (y=j)) \Vert$.

$\bullet$ Bounding $\Vert \mathbf{w}^{t} - \tilde{\mathbf{w}}^{t} \Vert$: 
We follow the identical sampling distribution (i.e., $\{ p_1, p_2, \cdots, p_{|\mathcal{K}|} \}$) to select $|\mathcal{S}_t|$ nodes from $|\mathcal{K}|$ nodes and let $\mathcal{S}_t = \{k_1, \cdots, k_{|\mathcal{S}_t|} \}$ denote the set of indices of chosen nodes. The global model in FL with partial node participation is represented as $ \mathbf{w}^{t} = \frac{1}{|\mathcal{S}_t|} \sum_{i=1}^{|\mathcal{S}_t|} \mathbf{w}_{k_i}^{t}$. Taking expectation over $\mathcal{S}_t$, we have
\begin{align*}
\label{C7}
\mathbb{E}_{\mathcal{S}_t} \Vert \mathbf{w}^{t} - \tilde{\mathbf{w}}^{t} \Vert = \mathbb{E}_{\mathcal{S}_t} \frac{1}{|\mathcal{S}_t|} \sum_{i=1}^{|\mathcal{S}_t|} \Vert \mathbf{w}_{k_i}^{t} - \tilde{\mathbf{w}}^{t} \Vert = \sum_{i=1}^{|\mathcal{K}|} p_i \Vert \mathbf{w}_i^{t} - \tilde{\mathbf{w}}^{t}  \Vert, \tag{B7}
\end{align*}
where the last equality in (\ref{C7}) is obtained by the following the observation $\mathbb{E}_{\mathcal{S}_t} \sum_{i\in \mathcal{S}_t} x_i = \mathbb{E}_{\mathcal{S}_t} \sum_{i=1}^{|\mathcal{S}_t|} x_{k_i} = |\mathcal{S}_t|  \mathbb{E}_{\mathcal{S}_t}x_{k_i} = |\mathcal{S}_t|\sum_{i=1}^{|\mathcal{K}|}  p_ix_i$ given $\mathcal{S}_t = \{x_{k_1}, \cdots, x_{k_{|\mathcal{S}_t|}} \} \subset \mathcal{K}$, and by replacing $x_i$ with $\mathbf{w}_i^{t}$ in the above observation.

We consider the model parameter in previous global round $\mathbf{w}_{i}^{t-1}$, which is identical for any $i\in \mathcal{K}$. As such, we have $\sum_{i=1}^{|\mathcal{K}|}  p_i ( \mathbf{w}_i^{t} - \mathbf{w}^{t-1}) = \tilde{\mathbf{w}}^{t}  - \tilde{\mathbf{w}}^{t-1}$. Thus, the above equation can be bounded as
\begin{align*}
\label{C8}
\sum_{i=1}^{|\mathcal{K}|} p_i \Vert \mathbf{w}_i^{t} - \tilde{\mathbf{w}}^{t}  \Vert & = \sum_{i=1}^{|\mathcal{K}|} p_i \Vert ( \underbrace{\mathbf{w}_i^{t} - \tilde{\mathbf{w}}^{t-1}}_{\mathbf{X}}) - (\tilde{\mathbf{w}}^{t} -\tilde{\mathbf{w}}^{t-1}) \Vert  \\
& \leq \sum_{i=1}^{|\mathcal{K}|} p_i \Vert \mathbf{w}_i^{t} - \tilde{\mathbf{w}}^{t-1} \Vert, \tag{B8}
\end{align*}
where the last equality holds because $\mathbb{E}  \Vert \mathbf{X} - \mathbb{E}[\mathbf{X}] \Vert \leq  \mathbb{E}  \Vert \mathbf{X} \Vert $. 

Substituting (\ref{C8}) into (\ref{C7}), we have,
\begin{align*}
\label{C9}
\mathbb{E}_{\mathcal{S}_t} \Vert \mathbf{w}^{t} - \tilde{\mathbf{w}}^{t} \Vert & \leq \sum_{i=1}^{|\mathcal{K}|} p_i\Vert \mathbf{w}_i^{t} - \tilde{\mathbf{w}}^{t-1} \Vert \\
& \leq \sum_{i=1}^{|\mathcal{K}|} p_i \Vert \mathbf{w}_i^{t} - \mathbf{w}_i^{t-1} \Vert \\
& \leq \sum_{i=1}^{|\mathcal{K}|} p_i  \Vert  \eta \nabla F_i(\mathbf{w}^{t-1}) \Vert \\
& \leq \eta \sum_{i=1}^{|\mathcal{K}|}  p_i\gamma_i, \tag{B9}
\end{align*}
where the last inequality results from Assumption \ref{ass3}.  

Finally, Theorem \ref{thm2} is proved by substituting (\ref{C9}) and (\ref{C6}) into (\ref{C1}). 

\bibliographystyle{ieeetr}
\bibliography{ref}

\end{document}